%% file: main.tex
\documentclass[letterpaper]{article} 
\usepackage{aaai2027}  
\usepackage{newtxmath}
\usepackage[hyphens]{url}  
\usepackage{graphicx} 
\usepackage{natbib}  
\usepackage{caption} 
\usepackage{booktabs}
\usepackage{amsmath}
\usepackage{multirow}
\usepackage{array}
\usepackage{siunitx}
\usepackage[capitalize,noabbrev]{cleveref}
\newcommand{\clade}{\textsc{Clade}}
\title{\clade{}: Category-Level Answering from Deficient Evidence\\in Retrieval-Augmented Language Models}
\author{S M Rayeed, Charles V. Stewart, Yao Ma}
\affiliations{Rensselaer Polytechnic Institute, Troy, NY, USA\quad \{rayees, stewart, may13\}@rpi.edu}
\begin{document}

\maketitle

\input{Sections/0.abstract}

\input{Sections/1.intro}

\input{Sections/2.related_work}

\input{Sections/3.data_setting}

\input{Sections/4.methodology}

\input{Sections/5.results}

\input{Sections/6.dislimcon}

\bibliography{main}

\clearpage

\appendix

\begin{center}
{\Large\bfseries Supplementary Material}
\end{center}

\input{Supp_sections/0_contents}

\input{Supp_sections/A_overview}
\input{Supp_sections/B_data}
\input{Supp_sections/C_statistics}
\input{Supp_sections/D_questions}
\input{Supp_sections/E_edits}
\input{Supp_sections/F_fixes}
\input{Supp_sections/G_globi}
\input{Supp_sections/H_recovery}
\input{Supp_sections/I_cases}
\input{Supp_sections/J_compute}
\input{Supp_sections/K_followup}

\end{document}

%% file: Sections/0.abstract.tex
\begin{abstract}
Retrieval-augmented systems often retrieve passages that are accurate and relevant but state the answer at a coarser level of granularity than the question requires. For example, a question may ask which specific disease a drug treats (\emph{familial hyperlipidemia}), while the retrieved passage mentions the broader disease family (\emph{hyperlipidemia}). Prior work on context sufficiency treats such coarse passages simply as insufficient. We argue that they form a distinct case, which we call \emph{resolvable insufficiency}: text alone cannot settle the fine-grained answer, but the broader category it mentions maps to a short list of candidate answers in a domain ontology. When language models read these coarse, \emph{family-only} passages, they exhibit a phenomenon we call \clade{} (Category-Level Answering from Deficient Evidence): they answer instead of abstaining or searching further. On biomedical questions from PrimeKG, family-only passages reduce abstention by 13.8 points compared to uninformative passages, and the resulting guesses are wrong about half the time. A counterfactual test confirms the cause: replacing an exact disease name with its broader family name lowers abstention by 17.4 points compared to deleting the sentence. Simple fixes like confidence filtering, scaling to a larger model, and explicitly instructing the model to abstain cannot resolve \clade{}, and this replicates on ecological species interactions in GloBI. Finally, we test whether an ontology can turn coarse passages into exact answers. Given the correct family, an ontology list includes the true answer in 98.5\% of cases, but string matching finds that family on the page for only 19.8\%. Letting the model choose the family raises this to 30.0\% and end-to-end recovery from 10.0\% to 26.8\%. Overall, our results show when language models should stop guessing and consult an ontology, and highlight which steps of that process remain unsolved. Codes are available at: \url{https://anonymous.4open.science/r/CLADE}.
\end{abstract}

%% file: Sections/1.intro.tex
\section{Introduction}

\begin{figure*}[t]
\centering
\includegraphics[width=\textwidth]{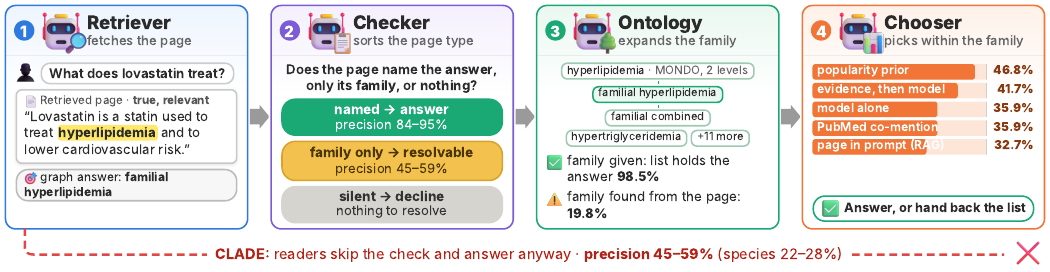}
\caption{\emph{Resolvable insufficiency} as an agent workflow, with the bottleneck we measure at each stage. The retrieved passage is accurate and relevant but names only the answer's family. A page-type check should send it to an ontology, which expands the family into a short candidate list for a chooser. The dashed red path is \clade{}: the model guesses from the family-only passage instead. In current practice, only the ontology step works reliably.}
\label{fig:concept}
\end{figure*}

Retrieval-augmented generation (RAG) systems answer questions by reading retrieved text \citep{lewis2020rag}. Recent work asks what a model should do when that text does not contain the complete answer. The usual framing is binary: a passage is either sufficient or insufficient \citep{joren2025sufficient}, and a reliable model answers in the first case and abstains in the second \citep{wen2025abstention}. In practice, many retrieved passages fall between these two extremes. Consider the question \emph{What disease does lovastatin treat?}, where the true answer is the specific condition \emph{familial hyperlipidemia}. Suppose the retriever returns a passage stating that \emph{lovastatin is used to treat hyperlipidemia} (~\cref{fig:concept}). This passage is accurate and relevant. It rules out every disease outside the hyperlipidemia family, but it is too coarse to specify which exact subtype is meant. In this case, a system can guess a specific disease, abstain, or look further using a medical ontology that lists specific members of the hyperlipidemia family.

We call this phenomenon \emph{resolvable insufficiency}. The passage is insufficient on its own because it states the answer at a coarse category level rather than the required fine-grained entity level. Yet it is resolvable because the coarse name maps directly to a short, finite set of candidate answers in a structured ontology. When a model instead guesses a fine-grained answer directly from such coarse text, we name the behavior \clade{} (Category-Level Answering from Deficient Evidence). Although this setting occurs frequently in real text with biomedical documents state the exact disease name for only a small fraction of knowledge-graph facts while describing many others at the family level, prior work overlooks how coarse evidence affects model abstention. Rather than proposing a surface-level patch, this paper digs into the mechanics of this phenomenon to understand how coarse evidence misleads large language models (LLMs) and where a neural reader should hand off to symbolic knowledge.

To study this systematically, we group retrieved passages into three categories based on how specifically they describe the target answer: \emph{named} (exact-match) passages, which explicitly mention the fine-grained target entity or an exact synonym; \emph{family-only} (coarse) passages, which omit the specific answer but mention a broader parent category (such as a disease family or biological genus) within two levels above the answer in an ontology; and \emph{silent} (unmentioned) passages, which discuss the query subject (e.g., the drug) but mention neither the specific answer nor its broader family. Using this framework, we investigate three research questions: (1) how often reference texts state fine-grained biomedical facts versus coarse categories (and whether a coarse name helps a model on its own), (2) whether family-only passages cause models to stop abstaining and guess inaccurately, and (3) whether a domain ontology can turn a coarse family mention into the exact fine-grained answer. Our main contributions are as follows:
\begin{itemize}
\item We define \emph{resolvable insufficiency} and \clade{}, and measure how text states 1,800 PrimeKG facts and 23,694 GloBI interactions. Text omits the exact disease for 77.5--84.9\% of PrimeKG facts, and a family name alone gives the model no closed-book advantage.
\item On 856 PrimeKG questions and three 7--8B models, family-only passages cut abstention by 13.8 points relative to silent passages. About half of the extra answers are wrong, and precision falls from 84--95\% to 45--59\%.
\item Counterfactual edits show that the family name causes this drop ($-17.4$ points against deleting the sentence). Confidence filtering, a 14B model and an instruction to abstain do not stop \clade{}, and the effect replicates on 900 GloBI questions. In free text, no model gives the exact answer from a family-only passage.
\item On 410 family-only facts, an ontology list holds the answer 98.5\% of the time once the family is known, but string matching finds the family from the passage only 19.8\% of the time. A model that chooses the family raises this to 30.0\% and end-to-end recovery from 10.0\% to 26.8\%, and a popularity prior beats the model at choosing within the list.
\end{itemize}

%% file: Sections/2.related_work.tex
\section{Related Work}

\paragraph{Insufficient context.}
Retrieval failures are a common weakness in RAG systems \citep{barnett2024seven}. \citet{joren2025sufficient} study the errors through the lens of context sufficiency. They label retrieved passages as either sufficient or insufficient and find that LLMs often hallucinate instead of abstaining when context is insufficient. Although they note that a passage can narrow down the answer without fully settling it, they group all such cases under insufficient context. Prior work also shows that models are easily thrown off by irrelevant text \citep{shi2023distracted,yoran2024robust}, and \citet{setiawan2026misleading} show that misleading passages cause prompt-based abstention to fail, whereas missing passages do not. Our focus, \emph{resolvable insufficiency}, differs from both irrelevant and misleading context. A coarse passage is factually accurate and points in the right direction, yet this partial relevance is enough to trigger \clade{} and break abstention.

\paragraph{Partial evidence.}
To handle imperfect passages, several methods train or prompt models to grade evidence quality. Self-RAG \citep{asai2024selfrag} trains a model to judge whether a passage fully or partially supports a generated answer, and ALCE \citep{gao2023alce} checks whether citations back up model claims. Similarly, CRAG \citep{yan2024crag} scores retrieved documents to trigger new searches when support is weak, and SURE-RAG \citep{qiu2026sure} checks whether context is sufficient before answering. While these methods detect that support is partial, they do not identify what kind of information is missing. A coarse mention is a structured form of partial evidence: it shows that fine-grained entity detail is missing and indicates where to find the candidate answers in an ontology.

\paragraph{Abstention.}
When evidence falls short and cannot be resolved, a reliable model should abstain. Recent surveys and benchmarks show that models often guess when they should hold back \citep{wen2025abstention,kirichenko2025abstentionbench}, including in clinical settings \citep{machcha2026abstain}. Most existing methods focus on detecting when a model lacks knowledge so it can decline to answer \citep{kadavath2022know,feng2024abstain}. While those studies measure whether a model abstains overall, we test how the granularity of retrieved evidence (\emph{named}, \emph{family-only}, \emph{silent}) causes a model to stop abstaining and guess.

\paragraph{Ontologies in RAG.}
Rather than guessing or abstaining on a coarse passage, a system can consult a structured ontology. Recent RAG systems use symbolic structures to guide retrieval: OG-RAG \citep{sharma2024ograg} grounds retrieval in domain ontologies, and TaSR-RAG \citep{sun2026tasr} uses taxonomies for multi-hop retrieval. Broader work on combining LLMs with knowledge graphs and symbolic reasoning is surveyed by \citet{pan2024unifying} and \citet{garcez2023third}. Unlike pipelines that query an ontology on every question, we study the specific case where a coarse passage calls for an ontology lookup. We also measure each step of the pipeline: extracting the family, listing its members, and picking the final answer; then we check where the process breaks down.

\paragraph{Fact coverage in text.}
Our analysis also connects to work on what reference and pretraining texts actually state. \citet{mallen2023trust} and \citet{kandpal2023longtail} link a model's closed-book knowledge to entity popularity and document counts, while LMEnt \citep{gottesman2026lment} tracks entity mentions across pretraining data. Whereas those studies count mentions of individual entities, we measure how often reference texts state relational facts at an exact (\emph{named}) level versus a coarse (\emph{family-only}) level.

\paragraph{Fine-grained biology.}
Biology shows the same gap between coarse and fine answers. On Bio-VQA, taxonomic accuracy falls from 0.72 at phylum to 0.12 at species \citep{rayeed2027biovqa}. And models lose accuracy from genus to species \citep{rayeed2025beetleverse} even on curated specimen images with validated traits \citep{rayeed2026beetles}. Knowledge-based VQA benchmarks also contain answers that their corpus does not support \citep{ma2026pitfalls}. We investigate the text side of this gap: a page that names the genus or family but not the answer.

%% file: Sections/3.data_setting.tex
\section{Data and Problem Setup}

\paragraph{Knowledge graphs and ontologies.}
We use drug-disease relations from PrimeKG, linked to the MONDO Disease Ontology. We reconstruct MONDO into 17,801 edges across 11,788 diseases (excluding parents with $>$\num{400} descendants). To test generalizability beyond medicine, we evaluate species interactions from GloBI, using 457,386 species-to-species edges across four relations (\emph{eats}, \emph{preys on}, \emph{parasite of}, \emph{pollinates}) with GloBI's taxonomy (species, genus, family) as the hierarchy.~\cref{tab:data} summarizes both domains.

\input{tables/tab_data}

\paragraph{Reference text corpora.}
For PrimeKG drugs, we use English Wikipedia articles and PubMed abstracts. For GloBI organisms, we use species' Wikipedia articles (dropping stubs $<$\num{3000} characters). All articles are stripped of markup, templates, infoboxes, tables, and file links.

\paragraph{Evidence granularity.}
We classify retrieved text into three levels: \emph{named} (exact name, synonym, abbreviation), \emph{family-only} (broad parent category within two levels), and \emph{silent} (neither). A control with unrelated disease pairs shows that matches track facts, not the size of the name set (unrelated matches occur in only 0.9--1.4\% of Wikipedia articles, versus 6.7--19.3\% for true diseases).

\paragraph{Corpus coverage.}
Evaluating 1,800 PrimeKG facts with Qwen2.5 0.5B (closed book) reveals that for 66.4\% of facts, the text lacks the exact disease name and the model lacks the fact. As detailed in Table~\ref{tab:boxes} and Table~\ref{tab:coverage}, the exact disease is missing in 84.9\% of Wikipedia articles, 81.1\% of PubMed sets, and 77.5\% combined, with random guessing accounting for most closed-book correct answers on unnamed facts.

\input{tables/tab_boxes}
\input{tables/tab_coverage}

\paragraph{Closed-book recall.}
Although family-only text covers 13.3\% of PrimeKG facts, testing on Qwen2.5-7B shows that coarse family mentions do not reflect stronger underlying model knowledge: it answers 39.6\% of family-only facts, 32.0\% of silent facts and 65.7\% of named facts correctly (Appendix~B). Odds ratio (OR) compares the odds of a correct closed-book answer in two groups. OR$=$1 means no difference, and OR$<$1 means the first group is answered correctly less often. We pool the relations with the Mantel--Haenszel estimator; $p$ is the chance of a gap this large if the groups did not differ. Family-only facts are answered far less often than named facts ($\mathrm{OR}=0.27$, $p = 9\times10^{-9}$) and as often as silent facts ($\mathrm{OR}=1.10$, $p = 0.61$). While an exact Wikipedia mention yields a 3.7$\times$ increase in the odds of model knowledge, a family name provides no measurable advantage on its own. This motivates testing how models handle coarse partial evidence at inference time, and whether symbolic ontologies can recover exact answers.

%% file: tables/tab_data.tex
\begin{table}[t]
\centering
\small
\setlength{\tabcolsep}{3.5pt}
\begin{tabular}{@{}lrr@{}}
\toprule
 & PrimeKG & GloBI \\
\midrule
Source records & 8,100,498 & 7,822,670 \\
\midrule
\multirow{4}{*}{Edges} & indication 9,388 & eats 235,960 \\
 & contraind.\ 30,675 & preys on 87,678 \\
 & off-label 2,568 & parasite of 117,894 \\
 & & pollinates 15,854 \\
\midrule
Hierarchy & MONDO & GloBI taxonomy \\
\quad child--parent links & 17,801 & 77,852 \\
\quad families, $\ge$2 members & 3,182 & 8,370 \\
\quad median / max size & 3 / 150 & 3 / 315 \\
\midrule
Text & Wiki.\ + PubMed & Wikipedia \\
\quad sources with text & 574 drugs & 5,425 species \\
\quad facts measured & 3,444 & 23,694 \\
Questions & 856 & 900 \\
\bottomrule
\end{tabular}
\caption{Data for both domains. PrimeKG links drugs to diseases, and MONDO groups diseases into families. GloBI links species to species, and its taxonomy groups species into genera. A \emph{family} is a parent with two or more children. \emph{Facts measured} are edges we check for a mention of the answer.}
\label{tab:data}
\end{table}

%% file: tables/tab_boxes.tex
\begin{table}[t]
\centering
\small
\setlength{\tabcolsep}{4pt}
\begin{tabular}{@{}lrrrrr@{}}
\toprule
 & & \multicolumn{2}{c}{Text names it} & \multicolumn{2}{c}{Text does not} \\
\cmidrule(lr){3-4}\cmidrule(l){5-6}
Relation & $n$ & knows & not & knows & not \\
\midrule
Indication & 600 & 9.7 & 7.2 & 21.8 & 61.3 \\
Contraind. & 600 & 1.5 & 5.2 & 20.7 & 72.7 \\
Off-label & 600 & 4.3 & 6.2 & 24.2 & 65.3 \\
\midrule
All & 1,800 & 5.2 & 6.2 & 22.2 & \textbf{66.4} \\
\bottomrule
\end{tabular}
\caption{Share of PrimeKG facts (\%) by whether the drug's Wikipedia article
names the disease and whether Qwen2.5 0.5B picks it from five options closed
book (``knows'').}
\label{tab:boxes}
\end{table}

%% file: tables/tab_coverage.tex
\begin{table}[t]
\centering
\small
\setlength{\tabcolsep}{4pt}
\begin{tabular}{@{}llrrrr@{}}
\toprule
Relation & Text & $n$ & Named & Family & Silent \\
\midrule
\multirow{3}{*}{Indication} & Wikipedia & 1,722 & 20.3 & 15.5 & 64.2 \\
 & PubMed & 1,722 & 26.3 & 18.6 & 55.1 \\
 & both & 1,722 & 29.0 & 19.7 & 51.3 \\
\midrule
\multirow{3}{*}{Contraind.} & Wikipedia & 1,722 & 9.8 & 4.1 & 86.1 \\
 & PubMed & 1,722 & 11.4 & 5.4 & 83.2 \\
 & both & 1,722 & 15.9 & 6.9 & 77.2 \\
\midrule
PrimeKG & both & 3,444 & 22.5 & 13.3 & 64.2 \\
\midrule
Eats & Wikipedia & 8,000 & 4.0 & 6.6 & 89.3 \\
Preys on & Wikipedia & 8,000 & 2.2 & 4.4 & 93.4 \\
Parasite of & Wikipedia & 2,389 & 12.5 & 15.5 & 72.0 \\
Pollinates & Wikipedia & 5,305 & 2.2 & 8.0 & 89.7 \\
\midrule
GloBI & Wikipedia & 23,694 & 3.9 & 7.1 & 89.1 \\
\bottomrule
\end{tabular}
\caption{Share of knowledge-graph facts whose text names the answer, names only its family, or names neither. PrimeKG uses 574 drugs with three facts of each relation.}
\label{tab:coverage}
\end{table}

%% file: Sections/4.methodology.tex
\section{Methodology}

We run four experiments. We measure how family-only pages change abstention, edit pages to isolate the effect of the family name, test three common fixes, and test whether an ontology can recover the exact answer.

\subsection{Abstention Evaluation Setup}
\label{subsec:abstention}

\paragraph{Questions.}
We curate 856 multiple-choice questions from the PrimeKG indications and contraindications: 300 with a \emph{named} page, 256 with a \emph{family-only} page, and 300 with a \emph{silent} page. Each relation and group has 150 questions, except family-only contraindications, where only 106 facts exist. Each question has five candidate diseases and a sixth option, \emph{Not enough information to tell}. The four wrong options come from the answer's MONDO neighborhood: siblings first, then cousins, then random diseases only if both run out. We exclude the answer's ancestors and descendants and any other disease the drug has under the same relation. 

\paragraph{Pages.}
We ask each question twice: closed book and with a page. The page starts with the drug's cleaned Wikipedia article, followed by its PubMed abstracts. We keep the first 800 characters and a 280 character window around each mention of the answer or its family, merge overlapping windows, and cap the page at \num{1800} characters. A silent page has no mentions, so it is the article's opening.

\paragraph{Models and Scoring.}
We evaluate five instruction-tuned open-weight models from three families: Qwen2.5 0.5B, 7B and 14B \citep{qwen2025qwen25}, OLMo~2 7B \citep{olmo2024olmo2}, and Granite~3.1 8B \citep{granite2024}. The model sees the page, the question and options A--F. We take the letter with the highest next-token probability, renormalized over the six letters. Choosing \emph{Not enough information to tell} counts as a decline. As a format check, we also ask 120 matched questions with no options and score each generated answer by name matching.

\paragraph{Controls.}
Two differences between groups could bias the comparison. First, family-only answers more often have enough relatives to fill all four wrong options, which makes those questions harder. Our main analysis therefore keeps only questions whose four wrong options all come from the neighborhood: 106 named, 215 family-only and 160 silent. Appendix~D reports all 856 questions, with the same pattern.

Second, the groups hold different facts, which may differ in difficulty before any page is read. We therefore compare each question with itself. For a question $q$ and an outcome such as a decline, let $y_q^{\text{page}}=1$ if the outcome occurs when the model reads the page and $0$ otherwise, and let $y_q^{\text{closed}}$ be the same without the page. The \emph{page effect} of $q$ is $y_q^{\text{page}}-y_q^{\text{closed}}$. Our main measure is the difference in mean page effect between the family-only questions $F$ and the silent questions $S$:
{\setlength{\abovedisplayskip}{3pt}\setlength{\belowdisplayskip}{3pt}%
\begin{equation}
\Delta = \frac{1}{|F|}\sum_{q\in F}\bigl(y_q^{\text{page}}-y_q^{\text{closed}}\bigr) - \frac{1}{|S|}\sum_{q\in S}\bigl(y_q^{\text{page}}-y_q^{\text{closed}}\bigr).
\label{eq:did}
\end{equation}}
For declines, $\Delta<0$ means a family-only page pushes the model toward declining less than a silent page does. To pool the 7--8B models, we average each question's page effect over the models first, so each question counts once.

\paragraph{Statistics.}
Effects are in percentage points. Brackets give 95\% confidence intervals (CI) from 5,000 bootstrap resamples of questions; an interval that excludes \num{0} marks a reliable effect. $p$-values come from two-sided $z$-tests (Appendix~C).

\subsection{Counterfactual Evaluation}

\paragraph{Page edits.}
The groups in~\cref{subsec:abstention} test different facts. To hold the fact fixed, we take 130 named questions whose answer has a MONDO parent. Then we edit each page in two ways: \emph{family edit} replaces every mention of the answer with its parent's name, and \emph{removed edit} deletes every sentence that mentions the answer (\cref{fig:edit}). The question, options and model stay the same.

\paragraph{Simple fixes fail.}
We test three ways to curb \clade{}. \emph{Confidence filtering} uses the renormalized probability of the chosen option; we report the AUROC for separating correct from wrong answers. \emph{Scaling} compares Qwen2.5 at 0.5B, 7B and 14B. \emph{Instruction} adds a line to the prompt telling the model to choose \emph{Not enough information to tell} when the page names only a broad category.

\paragraph{Second domain.}
We repeat the main experiment and counterfactual edits on 900 GloBI questions: 75 per page type in each of four relations. Wrong options come from the same genus first, then the same family.

\subsection{Ontology Recovery}

We test recovery on 410 PrimeKG facts whose Wikipedia and PubMed text names only the disease family (Figure~\ref{fig:concept}). A drug often has several true diseases in a family (2.47 per list on average), so a pick counts as correct if it matches any of them. Recovery has three stages:

\begin{enumerate}
\item \textbf{Build the list.} The oracle \emph{given-family} list takes the narrowest MONDO ancestor named in the text and lists its descendants up to two levels down, capped at 64. The automatic \emph{page-built} list takes every MONDO parent named in the drug's Wikipedia article and joins their descendants; lists over 64 count as too broad. The \emph{model-chosen} list asks Qwen2.5-7B which of the families on the page (up to ten) holds the answer and expands only that one.
\item \textbf{Choose one.} We compare five options: (1) Qwen2.5-7B closed book, (2) Qwen2.5-7B with the page in the prompt, (3) a reverse lookup to check whether the candidate's own Wikipedia article names the drug, (4) PubMed co-mention counts with relation keywords, raw and normalized by the disease's PubMed volume, and (5) a \emph{popularity prior} that picks the disease with the most PubMed abstracts, ignoring the drug.
\item \textbf{Compare without ontology.} Two baselines replace the family list with a list of the same length that holds the answer plus random diseases, scored by the model with or without the page. We test paired differences with McNemar's test \citep{mcnemar1947}.
\end{enumerate}

%% file: Sections/5.results.tex
\section{Results and Analysis}

\subsection{Coarse evidence weakens abstention (\clade{})}

\input{tables/tab_outcomes}

\begin{figure}[t]
\centering
\includegraphics[width=\columnwidth]{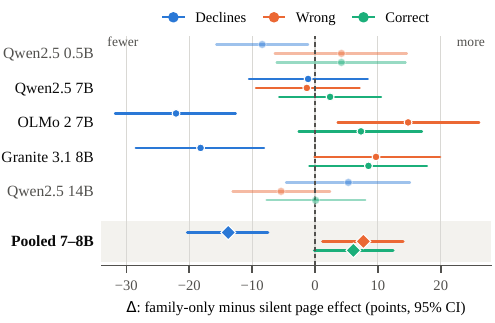}
\caption{Effect of a family-only page relative to a silent page. Each point is $\Delta$ from Eq.~\ref{eq:did} for one outcome, with its 95\% bootstrap CI; points left of zero mean the family-only page lowers that outcome. The pooled row averages the three 7-8B models over 481 questions. Faded rows are models outside the pooled set.}
\label{fig:forest}
\end{figure}

\paragraph{Models abstain less with partial evidence.}
\Cref{tab:outcomes} shows outcomes when the model reads the page. All 7-8B models abstain much less on a family-only page than on a silent page: Qwen2.5-7B from 83.8\% to 55.3\%, OLMo~2 from 45.0\% to 18.1\%, and Granite from 75.0\% to 43.3\%. Many of the extra answers are wrong. Precision, the share of answers that are correct, falls from 83.7-95.0\% on named pages to 45.5-59.4\% on family-only pages.

\paragraph{\clade{} drop comes from context.}
Family-only facts are harder even closed book: they draw more wrong guesses than silent facts in all three 7-8B models (Appendix~D). The page effect in Eq.~\ref{eq:did} removes this difference. Pooled over the models, a silent page raises abstention by $28.5$ points over closed book, but a family-only page raises them by only 14.7 points. The difference $\Delta$ is $-13.8$ points [$-20.3$, $-7.5$], and it stays significant after a Bonferroni correction for three outcomes ($p$=1.1$\times10^{-4}$; \cref{fig:forest}). The freed answers split between wrong answers ($+7.7$ [$1.2$, $14.0$]) and correct ones ($+6.1$ [$-0.1$, $12.4$]). In plain terms, a silent page makes the models stop and say they cannot tell; a family-only page makes them stop much less often, and they guess instead. The drop is significant for OLMo~2 ($-22.1$, $p$=6$\times10^{-6}$), Granite ($-18.2$, $p$=6$\times10^{-4}$) and Qwen2.5-0.5B ($-8.4$, $p$=$0.027$). Qwen2.5-7B shows almost no gap ($-1.1$, $p$=$0.82$). It already declines family-only questions far less often closed book (32.1\% vs.\ 59.4\%), and the page raises its abstention by about the same amount in both groups.

\paragraph{An example.}
For moxifloxacin, PrimeKG lists \emph{Klebsiella pneumonia}, and the page says only that the drug treats pneumonia. All five models pick some bacterial pneumonia with confidence up to 1.00, and none abstains. For lovastatin, OLMo~2 is right closed book but picks a sibling after reading a page that says only \emph{hyperlipidemia} (Appendix~I). Without options, failure looks different: in a free-text check on 120 questions, no model names the exact answer from a family-only page; instead 17.5-45.0\% answers give the family itself, which never occurs on named or silent pages (Appendix~K).

\paragraph{Family size.}
We split the family-only questions into thirds by family size. The drop is large for families of up to 10 members ($-17.3$ [$-26.2$, $-8.2$]) and of 11-20 members ($-18.4$ [$-26.6$, $-10.3$]). It shrinks to $-6.5$ [$-15.7$, $+3.0$] for families above 20 members (\cref{fig:fixes}c). A narrow family makes the page look more answerable.

\subsection{Counterfactual Edits confirm \clade{}}

\begin{figure}[t]
\centering
\includegraphics[width=\columnwidth]{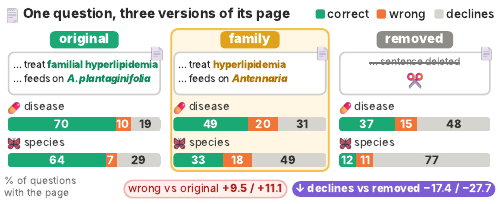}
\caption{Page edits on the same questions: PrimeKG (130) and GloBI (255). The \emph{family} edit replaces the answer with its parent; the \emph{removed} edit deletes the sentence. Bars show outcomes with the page (\%), averaged over the 7-8B models.}
\label{fig:edit}
\end{figure}

The edits change only the page (\cref{fig:edit}). Against the original page, the family edit lowers correct answers by 21.0 points and raises wrong answers by 9.5 points, while abstention rises by only 11.5 points. The key comparison is with the removed edit. Keeping the family name instead of deleting the sentence cuts abstention by 17.4 points [$-23.6$, $-11.5$] and raises wrong answers by 5.1 points [$0.5$, $10.3$]. Every model shows this order: from removed to family, abstention falls from 70.0\% to 43.8\% for Qwen2.5-7B, from 56.2\% to 40.8\% for Granite, and from 18.5\% to 7.7\% for OLMo~2. The family name alone suppresses abstention.

\subsection{Three Fixes Fall Short}

\begin{figure*}[t]
\centering
\includegraphics[width=\textwidth]{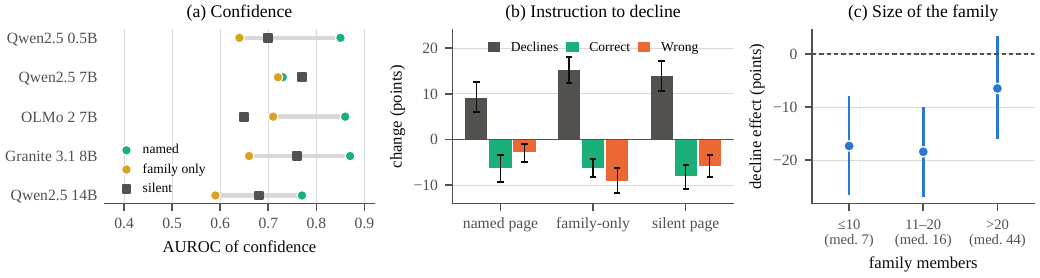}
\caption{Three candidate fixes. (a) Confidence separates correct from wrong answers poorly on family-only pages. (b) An instruction to decline on broad categories raises abstention on every page type, including named pages (pooled 7-8B models, 95\% CIs). (c) The drop in abstention is largest when the family has 20 or fewer members.}
\label{fig:fixes}
\end{figure*}

\paragraph{Confidence does not flag wrong answers.}
AUROC measures how well confidence ranks the correct answers above wrong ones: 1.0 is perfect and 0.5 is chance. On family-only pages it is 0.59-0.72, against 0.73-0.87 on named pages, in every model (\cref{fig:fixes}a). Wrong guesses are confident: their mean probability is 0.77-0.99 in 7-14B models. A threshold that keeps correct answers also keeps most wrong ones.

\paragraph{Scaling does not help.}
Qwen2.5-14B has the best precision on named pages (96.0\%). On family-only pages, precision rises from 38.1\% at 0.5B to 59.4\% at 7B, but then stops: 14B reaches 58.4\%, and it gives 14.7 points more wrong answers than on silent pages ($p$=6$\times10^{-6}$).

\paragraph{Instruction.}
Telling the model not to answer when the page names only a broad category raises abstention by 15.2 points on family-only pages (\cref{fig:fixes}b). It also raises them by 14.0 points on silent pages and 9.1 points on named pages, where it costs 6.3 points of correct answers. Precision on family-only pages stays at 47.9-64.9\%, far below named pages (89.6-94.7\%). The models scores change differently, OLMo~2 abstains far more on all page types, while Granite barely changes (Appendix~F). However, none of these models tells a family name from a specific answer.

\subsection{Second Domain: GloBI}

\input{tables/tab_globi}

Every finding on PrimeKG replicates on species interactions (\cref{tab:globi}). A silent page raises abstention by 12.1 points, but a page that names only the genus or family lowers them by 2.3 points ($\Delta$=$-14.3$, $p$=1.4$\times10^{-9}$). The effect is largest in OLMo~2 ($-31.7$). Here the extra answers are mostly wrong ($+11.6$ wrong, $p$=2.5$\times10^{-8}$; $+2.7$ correct, $p$=$0.09$). Precision on family-only pages falls to 22-28\%, close to the 20\% chance rate, because sibling species are harder to tell apart than sibling diseases. The page edit is even sharper: on 255 questions, keeping the genus name instead of deleting the sentence cuts abstention by 27.7 points [$-32.2$, $-23.5$] and raises wrong answers by 6.7 points (\cref{fig:edit}). \clade{} comes from hierarchical evidence in text, not from medical terms (Appendix~G).

\subsection{Ontology Recovery}

\begin{figure}[t]
\centering
\includegraphics[width=0.9\columnwidth]{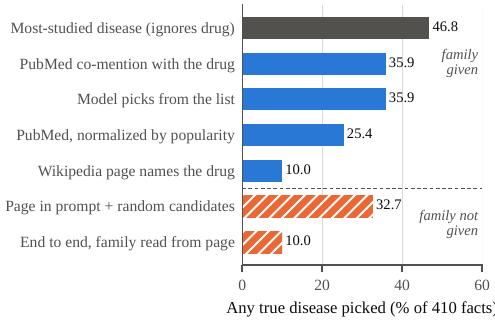}
\caption{Accuracy of choosing the exact disease for 410 family-only facts. Blue and gray bars choose from the given-family list; hatched bars do not receive the true family.}
\label{fig:recovery}
\end{figure}

\input{tables/tab_lists}

\paragraph{List coverage.}
When the right family is given, its MONDO descendants form a list of median 14 candidates. The list holds the sampled answer 98.5\% of the time and some true answer 99.5\% of the time (\cref{tab:lists}).

\paragraph{Finding the family.}
When the family must come from the drug's Wikipedia page, usable lists hold the answer for only 19.8\% of facts. The family is rarely missing; it is buried. A drug article names a median of five MONDO categories, such as comorbidities, side effects and other uses, and their union exceeds 64 candidates for 209 of 410 facts. PubMed makes this worse: coverage drops to 3.7\%. A model does better than string matching. When Qwen2.5-7B reads the page and picks one of the families it names, the chosen list holds the answer for 30.0\% of facts (44.9\% for any true answer), above the first-mentioned family (25.1\%) and a random one (13.9\%). Its lists are also short, with a median of 12 candidates. The ceiling is 50.0\%: for half of the facts, the right family is not among the usable families on the page.

\paragraph{Choosing a member.}
Within the given-family list (\cref{fig:recovery}), Qwen2.5-7B alone reaches 35.9\% (32.4\% with the page in the prompt), and raw PubMed co-mentions also reach 35.9\%. Evidence first, with the model as a fallback, reaches 41.7\%. Yet the best chooser ignores the drug: picking the disease with the most PubMed abstracts reaches 46.8\%. When we divide co-mentions by each disease's abstract count, accuracy drops from 35.9\% to 25.4\%, so co-mentions work mainly because common diseases are studied more.

\paragraph{Errors and end-to-end recovery.}
Most misses pick a better-studied sibling, such as \emph{ovarian carcinoma} instead of \emph{malignant germ cell tumor of ovary} for carboplatin, or a valid use missing from PrimeKG, so strict graph scoring understates accuracy (Appendix~H). Without the given family, page-built lists recover the answer for 10.0\% of facts with Wikipedia and 1.5\% with PubMed added; model-chosen lists recover 26.8\% (\cref{tab:lists}).

%% file: tables/tab_outcomes.tex
\begin{table}[t]
\centering
\small
\setlength{\tabcolsep}{2.9pt}
\begin{tabular}{llcccc}
\toprule
Model & Passage Type & Correct & Wrong & Declines & Prec. \\
\midrule
\multirow{3}{*}{Qwen2.5 7B} & named & 71.7 & 3.8 & 24.5 & 95.0 \\
 & family-only & 26.5 & 18.1 & 55.3 & 59.4 \\
 & silent & 13.8 & 2.5 & 83.8 & 84.6 \\
\midrule
\multirow{3}{*}{OLMo 2 7B} & named & 77.4 & 15.1 & 7.5 & 83.7 \\
 & family-only & 37.2 & 44.7 & 18.1 & 45.5 \\
 & silent & 30.6 & 24.4 & 45.0 & 55.7 \\
\midrule
\multirow{3}{*}{Granite 3.1 8B} & named & 72.6 & 7.5 & 19.8 & 90.6 \\
 & family-only & 30.2 & 26.5 & 43.3 & 53.3 \\
 & silent & 15.0 & 10.0 & 75.0 & 60.0 \\
\midrule
\multirow{3}{*}{Qwen2.5 0.5B} & named & 77.4 & 14.2 & 8.5 & 84.5 \\
 & family-only & 34.4 & 55.8 & 9.8 & 38.1 \\
 & silent & 28.7 & 50.6 & 20.6 & 36.2 \\
\midrule
\multirow{3}{*}{Qwen2.5 14B} & named & 67.9 & 2.8 & 29.2 & 96.0 \\
 & family-only & 24.2 & 17.2 & 58.6 & 58.4 \\
 & silent & 11.9 & 2.5 & 85.6 & 82.6 \\
\bottomrule
\end{tabular}
\caption{Open-book outcomes (\%) on PrimeKG questions where all four distractors come from the answer's MONDO neighborhood (106 named, 215 family-only, and 160 silent questions). Precision is the percentage of correct answers among questions where the model did not abstain.}
\label{tab:outcomes}
\end{table}

%% file: tables/tab_globi.tex
\begin{table}[t]
\centering
\footnotesize
\setlength{\tabcolsep}{2.6pt}
\begin{tabular}{lrr}
\toprule
Metric & PrimeKG & GloBI \\
\midrule
Questions in pooled set (all-neighbor) & 481 & 833 \\
Abstention page effect, family $-$ silent & $-13.8$ & $-14.3$ \\
Wrong-answer page effect, family $-$ silent & $+7.7$ & $+11.6$ \\
Precision on named passages (\%) & 84--95 & 84--95 \\
Precision on family-only passages (\%) & 45--59 & 22--28 \\
Confidence AUROC on named passages & .73--.87 & .82--.86 \\
Confidence AUROC on family-only passages & .66--.72 & .52--.68 \\
Edit: abstention, family $-$ removed & $-17.4$ & $-27.7$ \\
Edit: wrong answers, family $-$ removed & $+5.1$ & $+6.7$ \\
Edit: correct answers, family $-$ original & $-21.0$ & $-30.8$ \\
\bottomrule
\end{tabular}
\caption{Comparison of biomedical drug--disease facts (PrimeKG with MONDO) and ecological species interactions (GloBI with biological taxonomy) across the three 7--8B models. Effects are reported in percentage points; ranges span individual models. Every difference reported has a 95\% CI excluding zero.}
\label{tab:globi}
\end{table}

%% file: tables/tab_lists.tex
\begin{table}[t]
\centering
\small
\setlength{\tabcolsep}{3pt}
\begin{tabular}{@{}lccc@{}}
\toprule
Metric & Given & Page-built & Model-chosen \\
\midrule
Usable lists (of 410) & 410 & 166 & 375 \\
Holds sampled answer & 98.5\% & 19.8\% & 30.0\% \\
Holds any true answer & 99.5\% & 24.6\% & 44.9\% \\
Median list size & 14 & 32 & 12 \\
End-to-end recovery & -- & 10.0\% & 26.8\% \\
\bottomrule
\end{tabular}
\caption{Candidate lists for the 410 family-only facts: given family (oracle), all families on the page (page-built), and the one family Qwen2.5-7B picks from the page (model-chosen). Of the 244 unusable page-built lists, 209 exceed 64-candidate cap and 35 match no multi-child MONDO family. End-to-end recovery uses the best chooser for each list.}
\label{tab:lists}
\end{table}

%% file: Sections/6.dislimcon.tex
\section{Discussion}

\paragraph{Hand-off point.}
Our results mark a clear boundary between the language model and the ontology. On a family-only page, models take the family as license to guess. The ontology turns the same family into a short list that holds the answer, once the system knows which family on the page belongs to the question.

\paragraph{A three-step workflow.}
This suggests a simple workflow (\cref{fig:concept}). (1)~\textbf{Check the page type}, and answer from the page only if it names the answer. (2)~\textbf{Find the target family} tied to the queried relation; if none fits, abstain. (3)~\textbf{Expand and choose} with evidence and a popularity prior; answer only when one candidate clearly leads, otherwise return the list. Current models skip steps 1 and 2, and neither confidence nor instructions restore step~1. Step~2 is the main bottleneck: string matching finds the right family for 19.8\% of facts, and a model chooser raises this to 30\% against a ceiling of 50\%. Any chooser for step~3 should beat the popularity prior.

\paragraph{Beyond sufficiency.}
Recent work asks whether a page is sufficient \citep{joren2025sufficient}; our results show that \emph{how} it falls short matters. A family-only page differs from a silent one: it carries real signal, it breaks abstention, and it points into a symbolic hierarchy. Treating it as its own case lets a RAG system send partial evidence to an ontology instead of guessing or discarding it.

\section{Limitations}

We study two domains and run recovery only on PrimeKG. We use multiple-choice questions with an explicit decline option, which controls the wrong options and the scoring but may differ from free-text refusals. A free-text check points the same way (Appendix~K): no model gives the exact answer from a family-only page, and 17.5--45.0\% of its answers name the family itself. We test open models up to 14B, and the page edits cover only named facts whose answer has a parent. Page-built lists rely on exact string matching. Family-only contraindications are few (106 questions).

\section{Conclusion}

Retrieved text often states facts at a coarser level than the question asks. We show that this \emph{resolvable insufficiency} triggers \clade{} in medicine and ecology: a family or genus name leads models to stop declining and guess, which roughly halves precision. An ontology can narrow the answer to a short list that holds it 98.5\% of the time once the right family is found, and a model that chooses the family from the page already raises end-to-end recovery from 10.0\% to 26.8\%. Finding the family more reliably and beating a popularity prior remain the key open problems for neuro-symbolic RAG systems that must know when to hand off to an ontology.

%% file: Supp_sections/0_contents.tex
\section*{Contents}
\label{sec:supp-contents}

We organize this supplement into eleven sections, in the same order as the main paper. Every number comes from the same per-question outputs that produce the tables and figures of the main paper.

\begin{itemize}\itemsep0.4em

  \item[\textbf{A}] \textbf{Overview and reproducibility.} We summarize the four experiments in one figure. We describe the released code and data, and we give the commands that rerun every table without a GPU and the full pipeline on a cluster.

  \item[\textbf{B}] \textbf{Data and page types.} We describe how we build both graphs: PrimeKG with MONDO, and GloBI with its own taxonomy. We explain how we clean Wikipedia and query PubMed. We define the three name sets and the rule that sorts each page as named, family-only or silent. We then report the chance control, in which unrelated diseases match the same text only 0.9--1.4\% of the time, and the closed-book check, which shows that a family name does not mark an easier fact.

  \item[\textbf{C}] \textbf{Statistical procedures.} We define every statistic in the paper with an equation: the page effect, the difference in differences $\Delta$, pooling over models, the bootstrap intervals, the $z$-test and Bonferroni correction, the Mantel--Haenszel odds ratio, AUROC and McNemar's test.

  \item[\textbf{D}] \textbf{Abstention questions and full results.} We show how we sample the 856 questions, write each question, and draw its wrong options from the answer's MONDO neighborhood. We give the windowed page, the prompt, and the way we read the answer from letter probabilities. We then report the full results: closed-book outcomes, all 856 questions, per-model page effects, results by relation, and results by family size.

  \item[\textbf{E}] \textbf{Page edits.} We describe how we build the family and removed versions of each named page. We give per-model outcomes for all three versions in both domains, and the pooled edit effects with their intervals.

  \item[\textbf{F}] \textbf{Three fixes.} We give per-model confidence on right and wrong answers, with AUROC on each page type. We follow precision across the three Qwen2.5 sizes. We give the exact instruction sentence and show how differently each model responds to it.

  \item[\textbf{G}] \textbf{Second domain: GloBI.} We describe the GloBI page types, question templates and wrong options. We give per-model outcomes closed book and with the page, and per-model page effects, where OLMo~2 shows the largest drop ($-31.7$ points). We explain why precision on family-only pages falls close to chance.

  \item[\textbf{H}] \textbf{Ontology recovery.} We describe how we build the given-family and page-built lists, with pseudo-code, and why page-built lists fail. We define each chooser, including the combined evidence score and the popularity prior, and report how often each signal fires and how accurate it is. We compare the family lists with random lists of the same length, and we show examples of recovered and missed answers.

  \item[\textbf{I}] \textbf{Worked examples.} We follow single questions through the test, with real pages and each model's answer: a confident wrong guess for moxifloxacin, a page that makes one model worse for lovastatin, and three GloBI pages that are named, family-only and silent.

  \item[\textbf{J}] \textbf{Software and compute.} We list the model checkpoints and describe how we run them, batch them, cache their answers and fix every random seed.

  \item[\textbf{K}] \textbf{Choosing the family and free-text answers.} We test a first fix for step~2, in which Qwen2.5-7B picks the family on the page. It raises list coverage from 19.8\% to 30.0\% and end-to-end recovery from 10.0\% to 26.8\%. We then ask 120 questions without options. No model gives the exact answer from a family-only page, and many give the family name itself.

\end{itemize}

\paragraph{How to read the tables.}
We report outcomes as percentages of questions and effects in percentage points. Square brackets give 95\% bootstrap confidence intervals; an interval that excludes zero marks a reliable effect. \emph{Precision} is the share of correct answers among the questions a model answers. \emph{Pooled} results average the three 7-8B models, Qwen2.5-7B, OLMo~2 7B and Granite~3.1 8B, over the same questions.

\paragraph{A note on terms.}
A \emph{page} is the retrieved passage the model reads. A \emph{family} is a parent of the answer up to two levels above it: a MONDO disease family, or a genus or family in GloBI. We use \emph{abstain} and \emph{decline} for the same outcome: the model chooses \emph{Not enough information to tell}.

%% file: Supp_sections/A_overview.tex
\section{Overview and Reproducibility}
\label{app:overview}

\begin{figure}[t]
\centering
\includegraphics[width=\columnwidth]{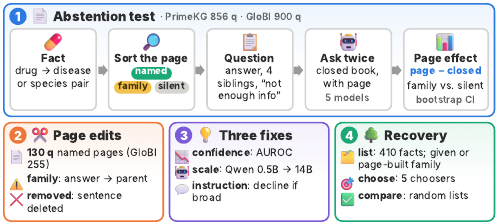}
\caption{Our four experiments. (1) We sort each page as named, family-only or silent, and ask each question closed book and with the page. (2) We edit named pages: the answer becomes its family name, or we delete the sentence. (3) We test three fixes. (4) We try to recover the exact answer from the ontology.}
\label{fig:method}
\end{figure}

This supplement gives the details behind each part of the main paper. It follows the same order as the paper: data and page types, statistics, the abstention test, page edits, the three fixes, the second domain, and ontology recovery. The last appendix covers two follow-up tests: a model that chooses the family on the page, and free-text answers. \Cref{tab:map} shows where each result of the main paper is explained in more depth. \Cref{fig:method} summarizes the four experiments in one picture.

\begin{table}[t]
\centering
\footnotesize
\setlength{\tabcolsep}{3pt}
\begin{tabular}{@{}p{3.9cm}p{3.6cm}@{}}
\toprule
Main paper & Supplement \\
\midrule
Data, page types, \cref{tab:data,tab:coverage} & \cref{app:data} \\
Odds ratios, CIs, $p$-values, Eq.~\ref{eq:did} & \cref{app:stats} \\
Questions, prompts, \cref{tab:outcomes}, \cref{fig:forest} & \cref{app:questions} \\
Page edits, \cref{fig:edit} & \cref{app:edits} \\
Confidence, scale, instruction, \cref{fig:fixes} & \cref{app:fixes} \\
GloBI, \cref{tab:globi} & \cref{app:globi} \\
Recovery, \cref{tab:lists}, \cref{fig:recovery} & \cref{app:recovery} \\
Worked examples & \cref{app:cases} \\
Software and compute & \cref{app:compute} \\
Choosing the family, free text & \cref{app:followup} \\
\bottomrule
\end{tabular}
\caption{Where each part of the main paper is explained in this supplement.}
\label{tab:map}
\end{table}

\paragraph{Code and data.}
We release all code and derived data at \url{https://anonymous.4open.science/r/CLADE}. The release holds the question sets, the windowed pages, the candidate lists, the per-question model outputs, and every script that builds them. We do not redistribute Wikipedia or PubMed text; the scripts download it again from the public APIs. PrimeKG, MONDO, GloBI, Wikipedia and PubMed are all public.

\paragraph{Rerunning the numbers.}
Every table and figure comes from the shipped per-question outputs, so a reader can check them without a GPU or network access:
\begin{quote}
\footnotesize\ttfamily
bash tools/check\_results.sh\\
\# reruns the analysis and compares it\\
\# with results/; prints "identical: 22"
\end{quote}
The full pipeline, from raw downloads to the final tables, runs with one command on a SLURM cluster:
\begin{quote}
\footnotesize\ttfamily
bash run\_all.sh~~~~~~~~~~~~~\# PrimeKG\\
bash globi.sh all~~~~~~~~~~~~\# GloBI
\end{quote}
Both scripts resume where they stop, and a smoke test on small synthetic data checks the whole pipeline in a few minutes (\texttt{--smoke}).

%% file: Supp_sections/B_data.tex
\section{Data and Page Types}
\label{app:data}

This appendix explains how we build the two graphs, how we read the text, and how we decide whether a page names the answer, names only its family, or says nothing.

\subsection{Graphs and Hierarchies}

\paragraph{PrimeKG and MONDO.}
PrimeKG has \num{8100498} rows. We keep three drug--disease relations: \num{9388} indications, \num{30675} contraindications and \num{2568} off-label uses. Each disease has a MONDO identifier. We rebuild MONDO as a table of child--parent links: \num{17801} links over \num{11788} diseases. A parent with two or more children is a \emph{family}; MONDO has \num{3182} of them. We drop parents with more than 400 descendants, such as \emph{disease} or \emph{disorder}. Naming them on a page tells the reader almost nothing, and they would turn almost every page into a family-only page.

\paragraph{GloBI and its taxonomy.}
We use the archived GloBI dump (Zenodo record 4460654), which holds \num{7822670} interaction records. We keep four relations: \emph{eats}, \emph{preys on}, \emph{parasite of} and \emph{pollinates}. Together they hold \num{3210597} records. We keep a record only when both sides are species, written as a two-word Latin name (``Genus epithet''). We drop names with ``sp.'', hybrids and numbers, because a question about a genus has no single answer. This leaves \num{457386} species-to-species edges:

\begin{center}
\footnotesize
\begin{tabular}{@{}lrr@{}}
\toprule
Relation & Edges & Source species \\
\midrule
eats & \num{235960} & \num{25637} \\
parasite of & \num{117894} & \num{31432} \\
preys on & \num{87678} & \num{5138} \\
pollinates & \num{15854} & \num{2818} \\
\bottomrule
\end{tabular}
\end{center}

GloBI records give a taxonomic path for each taxon (species, genus, family, order, class). We build the hierarchy from these paths. Different records can give a taxon different parents, so we keep the parent seen most often. This makes the table a tree: climbing it never moves sideways. The result has \num{77852} child--parent links, and \num{8370} genera have two or more species. As in MONDO, the median family has three members.

\subsection{Text Sources}

\paragraph{Wikipedia.}
For each drug, and for each consumer species in GloBI, we fetch the English Wikipedia article through the MediaWiki API. We turn the raw markup into plain prose. We remove comments, reference tags, templates (including infoboxes and citation blocks, with nested braces), tables, and file and category links. We replace each internal link with its visible text and remove bold and italic marks, HTML tags and character entities. We keep section headings as plain lines, so the text reads like the page a person sees. For GloBI we skip stubs shorter than \num{3000} characters.

\paragraph{PubMed.}
For each drug we retrieve PubMed abstracts through NCBI E-utilities. We split the query budget equally between a treatment query and an adverse-effect or contraindication query, so both relations get the same chance to appear. The drug's full text is its Wikipedia article followed by its abstracts.

\subsection{How a Page Names a Fact}

\paragraph{Three name sets.}
Each answer has three nested sets of names:
\begin{itemize}
\item the \emph{exact name}: its primary MONDO label, or its Latin name in GloBI;
\item \emph{alternative names}: titles of Wikipedia pages that redirect to the answer's article, which bring in synonyms, lay terms and abbreviations; in GloBI we also add the abbreviated form (\emph{A.~plantaginifolia});
\item \emph{family names}: the labels of the answer's ancestors up to two levels above it (the MONDO parent and grandparent, or the genus and family).
\end{itemize}

\paragraph{Page type.}
Let $T$ be the text, $N(a)$ the exact and alternative names of answer $a$, and $A_2(a)$ its family names. We assign each fact to the finest level the text matches:
\begin{equation}
\mathrm{type}(T,a)=
\begin{cases}
\text{named} & \text{if } T \text{ matches } N(a),\\
\text{family-only} & \text{else if } T \text{ matches } A_2(a),\\
\text{silent} & \text{otherwise.}
\end{cases}
\end{equation}

\paragraph{Matching rule.}
A name matches when it appears on word boundaries, ignoring case, with an optional trailing \emph{s} for simple plurals. A multi-word name matches across any run of white space. We ignore names shorter than four characters, because short acronyms match by accident.

\paragraph{Chance control.}
A family name is broader than a disease name, so it can match unrelated text more easily. To measure this, we pair each fact with three diseases from the same relation that have no link to the drug, and we run the same matcher on the same text. \Cref{tab:floor} compares the true disease with these unrelated ones on 1,800 PrimeKG facts in Wikipedia. The unrelated diseases match 0.9--1.4\% of the time at the named level and 1.8--3.3\% once family names are added. The true disease matches 6.7--19.3\% and 9.3--34.7\%. Matches therefore follow the fact, not the size of the name set.

\begin{table}[t]
\centering
\footnotesize
\setlength{\tabcolsep}{3.5pt}
\begin{tabular}{@{}lrrrr@{}}
\toprule
 & \multicolumn{2}{c}{True disease} & \multicolumn{2}{c}{Unrelated diseases} \\
\cmidrule(lr){2-3}\cmidrule(l){4-5}
Relation & named & + family & named & + family \\
\midrule
Indication & 19.3 & 34.7 & 1.4 & 2.7 \\
Contraindication & 6.7 & 9.3 & 0.9 & 1.8 \\
Off-label use & 11.8 & 23.7 & 0.9 & 3.3 \\
\bottomrule
\end{tabular}
\caption{Share of facts (\%) whose Wikipedia text matches the true disease or one of three unrelated diseases from the same relation (600 facts per relation; \num{1800} unrelated draws each). ``+ family'' also counts family names.}
\label{tab:floor}
\end{table}

\subsection{Text Coverage and Closed-Book Knowledge}

\paragraph{The four-box table.}
\Cref{tab:boxes} in the main paper uses 1,800 PrimeKG facts (600 per relation). A fact counts as \emph{text names it} when the drug's Wikipedia article contains the exact or an alternative name. A family-only page counts as \emph{text does not name it}. To test closed-book knowledge, Qwen2.5-0.5B scores the true disease against four random diseases from the same relation. It reads a short stem, such as ``\emph{X is a drug used to treat the disease called}'', and we take the mean log-probability of each disease name after it. The highest score is the model's pick.

\paragraph{Knowledge above chance.}
On facts the text does not name, the model picks the true disease 25.1\% of the time. With five options, chance is 20\%. We correct for guessing with
\begin{equation}
k=\frac{\mathrm{acc}-c}{1-c}=\frac{0.251-0.200}{1-0.200}\approx 6.4\%,
\end{equation}
where $c$ is the chance rate. So the model truly knows only about 6\% of these facts; the rest of its correct picks are lucky guesses.

\paragraph{Does a family name mean an easier fact?}
We test whether facts written at the family level are simply easier. Qwen2.5-7B answers the same 1,800 facts closed book (\cref{tab:closedbook}).

\input{tables/tab_closedbook}
 Within each relation we build a $2\times2$ table (family-only vs.\ named, or family-only vs.\ silent; correct vs.\ wrong) and pool the three tables with the Mantel--Haenszel odds ratio (\cref{app:stats}). Family-only facts are much harder than named facts ($\mathrm{OR}=0.27$ [0.17, 0.43]) and as hard as silent facts ($\mathrm{OR}=1.10$ [0.80, 1.53]). An exact name in the text goes with 3.7 times higher odds that the model knows the fact; a family name goes with none.

%% file: tables/tab_closedbook.tex
\begin{table}[t]
\centering
\small
\setlength{\tabcolsep}{4pt}
\begin{tabular}{@{}lrrr@{}}
\toprule
Relation & Named & Family only & Silent \\
\midrule
Indication & 84.2 {\scriptsize(101)} & 44.7 {\scriptsize(94)} & 44.0 {\scriptsize(405)} \\
Contraind. & 32.5 {\scriptsize(40)} & 18.8 {\scriptsize(16)} & 23.4 {\scriptsize(544)} \\
Off-label & 57.1 {\scriptsize(63)} & 37.5 {\scriptsize(72)} & 31.6 {\scriptsize(465)} \\
\midrule
All & 65.7 {\scriptsize(204)} & 39.6 {\scriptsize(182)} & 32.0 {\scriptsize(1,414)} \\
\bottomrule
\end{tabular}
\caption{Closed-book accuracy of Qwen2.5-7B by what the drug's Wikipedia page says about the answer. Chance is 20\%.}
\label{tab:closedbook}
\end{table}

%% file: Supp_sections/C_statistics.tex
\section{Statistical Procedures}
\label{app:stats}

This appendix defines every statistic in the paper. All tests are two-sided, and all code uses only the Python standard library, with random seed 0.

\paragraph{Page effect.}
For a question $q$ and an outcome $o\in\{\text{correct},\text{wrong},\text{decline}\}$, let $y_q^{o,\text{page}}=1$ if the model's answer has outcome $o$ when it reads the page, and $0$ otherwise. Let $y_q^{o,\text{closed}}$ be the same without the page. The page effect of $q$ is
\begin{equation}
d_q^{o}=y_q^{o,\text{page}}-y_q^{o,\text{closed}}\in\{-1,0,1\}.
\end{equation}
A page effect of $+1$ for declines means the page turned an answer into a decline; $-1$ means the page turned a decline into an answer. Our main measure (Eq.~\ref{eq:did}) is the mean page effect on family-only questions $F$ minus that on silent questions $S$:
\begin{equation}
\Delta^{o}=\bar d^{\,o}_F-\bar d^{\,o}_S,\qquad \bar d^{\,o}_G=\frac{1}{|G|}\sum_{q\in G}d_q^{o}.
\end{equation}
This is a difference in differences. The first difference (page minus closed book) removes how hard each fact is on its own. The second (family-only minus silent) asks what the family name adds beyond a page that says nothing.

\paragraph{Pooling models.}
For the pooled 7--8B result, we first average each question's page effect over Qwen2.5-7B, OLMo~2 7B and Granite~3.1 8B, and then compute $\Delta^o$ on these averages. Each of the 481 questions thus counts once, however many models answer it.

\paragraph{Confidence intervals.}
We draw 5,000 bootstrap samples. Each sample redraws the questions of each group with replacement, keeping the group sizes, and recomputes $\Delta^o$. The 95\% CI is the 2.5th to 97.5th percentile of these values. If the interval excludes zero, we call the effect reliable.

\paragraph{$p$-values.}
We test $\Delta^o=0$ with a Welch-type $z$-test, which allows the two groups to have different variances:
\begin{equation}
z=\frac{\bar d^{\,o}_F-\bar d^{\,o}_S}{\sqrt{s_F^2/|F|+s_S^2/|S|}},
\end{equation}
where $s_G^2$ is the sample variance of $d_q^o$ in group $G$. The $p$-value is $2\,(1-\Phi(|z|))$, with $\Phi$ the standard normal distribution function. A small $p$ means a gap this large is unlikely if the family name had no effect. Each comparison covers three outcomes, so the Bonferroni-corrected $p$-value is three times the raw one.

\paragraph{Odds ratios.}
For the closed-book check in \cref{app:data}, each relation $r$ gives a $2\times2$ table: $a_r$ and $b_r$ are correct and wrong answers in the first group, $c_r$ and $e_r$ in the second, and $n_r$ is the total. The Mantel--Haenszel odds ratio pools the relations:
\begin{equation}
\mathrm{OR}_{\mathrm{MH}}=\frac{\sum_r a_r e_r/n_r}{\sum_r b_r c_r/n_r}.
\end{equation}
$\mathrm{OR}=1$ means the two groups are answered correctly equally often. $\mathrm{OR}=0.27$ means the odds of a correct answer in the first group are about a quarter of those in the second. Pooling within relations keeps the result from being driven by one relation that happens to be easier.

\paragraph{AUROC.}
To judge confidence filtering, we take the answers a model gives (not its declines) and ask how well the probability of the chosen option ranks correct answers above wrong ones. AUROC is the chance that a random correct answer has higher confidence than a random wrong answer. An AUROC of 1.0 means a threshold can separate them perfectly; 0.5 means confidence carries no signal.

\paragraph{McNemar's test.}
In recovery, two choosers answer the same 410 facts. Let $b$ be the number of facts the first gets right and the second gets wrong, and $c$ the reverse. McNemar's test uses only these disagreements, with a continuity correction:
\begin{equation}
\chi^2=\frac{(|b-c|-1)^2}{b+c},
\end{equation}
compared with a $\chi^2$ distribution with one degree of freedom. It asks whether one chooser wins the disagreements more often than chance would allow.

%% file: Supp_sections/D_questions.tex
\section{Abstention Questions and Full Results}
\label{app:questions}

This appendix shows how we build each question, how the model sees it, and how we read its answer. It then gives the full results behind \cref{tab:outcomes} and \cref{fig:forest} of the main paper.

\subsection{Building the Questions}

\paragraph{Sampling.}
We sort PrimeKG facts into six cells: three page types (named, family-only, silent) times two relations (indication, contraindication). We shuffle each cell with seed 0 and take facts in order until the cell has 150 questions. We skip a fact when the drug has no text, when we cannot draw four wrong options, or when the 1,800-character page no longer shows the matched name or family. Only 106 family-only contraindications pass these checks, so we use all of them. This gives 856 questions.

\paragraph{Question templates.}
Indication questions ask: \emph{Which disease is X used to treat?} Contraindication questions ask: \emph{In patients with which disease must X not be given?}

\paragraph{Wrong options.}
Each question has five diseases and the option \emph{Not enough information to tell}, labeled A to F. We place the true disease and the decline option at random letters. We draw the four wrong options in this order:
\begin{enumerate}
\item \emph{siblings}: other children of the answer's MONDO parent;
\item \emph{cousins}: other grandchildren of the answer's grandparent;
\item \emph{random} diseases from the same relation, only if siblings and cousins run out.
\end{enumerate}
We never use an ancestor or descendant of the answer, or any other disease the drug has under the same relation in PrimeKG, since that option would also be correct. Because siblings and cousins share the answer's family, a page that names only the family cannot rule them out. Random diseases, however, can be ruled out by the family name alone, which makes a question easier. \Cref{tab:sib} shows that family-only questions more often have four neighborhood options than the other groups. This is why the main analysis keeps only questions whose four wrong options all come from the neighborhood (106 named, 215 family-only, 160 silent; 481 in all).

\begin{table}[t]
\centering
\footnotesize
\setlength{\tabcolsep}{4pt}
\begin{tabular}{@{}llrr@{}}
\toprule
Page type & Relation & $n$ & All-neighbor (\%) \\
\midrule
Named & both & 300 & 35.3 \\
Family-only & contraindication & 106 & 83.0 \\
Family-only & indication & 150 & 84.7 \\
Silent & contraindication & 150 & 45.3 \\
Silent & indication & 150 & 61.3 \\
\bottomrule
\end{tabular}
\caption{Questions per group and the share whose four wrong options are all siblings or cousins of the answer in MONDO.}
\label{tab:sib}
\end{table}

\paragraph{Windowed pages.}
A full drug article is too long for a prompt, so we cut a window around the evidence. The page starts with the first 800 characters of the drug's cleaned text. We then add a 280-character window on each side of every mention of the answer or its family. We merge windows that overlap, join them with ellipses, and stop at 1,800 characters. A silent page has no mentions, so it is just the opening of the article. Every model sees the same page for the same question.

\paragraph{Prompt.}
Each prompt is one user turn in the model's own chat template. The closed-book prompt is the same without the passage block:
\begin{quote}
\footnotesize\ttfamily
Passage:\\
<passage>\\[2pt]
Question: <question>\\
Options:\\
A. <option>\\
\ldots\\
F. <option>\\
Answer with the letter of one option only.
\end{quote}

\paragraph{Reading the answer.}
We do not generate text. We read the probability of the next token for each letter. For letter $\ell$, let $p(\ell)$ be the larger of the probabilities of ``$\ell$'' and ``\,$\ell$'' (with a leading space). We renormalize over the six letters:
\begin{equation}
\pi(\ell)=\frac{p(\ell)}{\sum_{m\in\{A,\dots,F\}}p(m)},
\end{equation}
and the model's answer is the letter with the highest $\pi$. Its confidence is that $\pi$. If the answer is \emph{Not enough information to tell}, the question counts as a decline. Precision is the share of correct answers among the questions the model does not decline.

\subsection{Full Results}

\paragraph{Without a page.}
\Cref{tab:closed_full} gives closed-book outcomes on the 481 matched questions. Family-only questions already draw more wrong answers and fewer declines than silent ones in every 7--8B model. For example, Qwen2.5-7B declines 32.1\% of family-only questions and 59.4\% of silent ones before it reads anything. If we compared pages directly, part of the gap would come from these facts, not from the page. The page effect removes this part.

\input{tables/tab_closed_full}

\paragraph{With a page, all questions.}
\Cref{tab:all_page} repeats \cref{tab:outcomes} on all 856 questions, without the neighborhood filter. The pattern is the same: family-only pages draw far more wrong answers than silent pages, with lower precision. Compared directly with the page, family-only pages give more wrong answers than silent pages by 15.6, 20.3 and 16.5 points for Qwen2.5-7B, OLMo~2 and Granite ($p=3\times10^{-6}$, $5\times10^{-5}$, $6\times10^{-5}$), and fewer declines by 28.4, 26.9 and 31.7 points ($p<10^{-7}$ each).

\input{tables/tab_all_page}

\paragraph{Page effects per model.}
\Cref{tab:did_full} breaks the page effect into its parts. For each outcome it shows the mean page effect on family-only questions, the same on silent questions, and their difference with its CI and $p$-value. The pooled row is the result in the main paper. OLMo~2 and Granite each show a large drop in declines; Qwen2.5-7B shows none, because it already declines family-only questions much less often closed book (\cref{tab:closed_full}).

\input{tables/tab_did_full}

\paragraph{By relation.}
\Cref{tab:byrel} splits wrong answers with the page by relation. Family-only pages give more wrong answers than silent pages for both relations in every 7--8B model. For indications, Qwen2.5-7B is wrong on 26.0\% of family-only questions and 2.2\% of silent ones. Models are more careful on contraindications and decline more of them, but the gap stays. Qwen2.5-0.5B rarely declines at all (8--21\% with a page, under 4\% closed book), so its wrong-answer rates are high everywhere.

\input{tables/tab_byrel}

\paragraph{Family size.}
\Cref{tab:famsize} splits the family-only questions into thirds by the size of the named family. The pooled drop in declines is about 17--18 points for families of up to 20 members and only 6.5 points, not significant, for larger families. A short list looks answerable to the model; a long one does not.

\begin{table}[t]
\centering
\footnotesize
\setlength{\tabcolsep}{3pt}
\begin{tabular}{@{}lrrlr@{}}
\toprule
Family size & $n$ & Median & Decline effect [95\% CI] & $p$ \\
\midrule
$\le 10$ & 71 & 7 & $-17.3$ [$-26.2$, $-8.2$] & .001 \\
11--20 & 69 & 16 & $-18.4$ [$-26.6$, $-10.3$] & $<$.001 \\
$>20$ & 68 & 44 & $-6.5$ [$-15.7$, $+3.0$] & .174 \\
\bottomrule
\end{tabular}
\caption{Decline effect ($\Delta$ for declines, points) by the number of members in the named family, pooled over the three 7--8B models.}
\label{tab:famsize}
\end{table}

%% file: tables/tab_closed_full.tex
\begin{table}[ht]
\centering
\small
\setlength{\tabcolsep}{2.8pt}
\begin{tabular}{llccc}
\toprule
Model & Group & Correct & Wrong & Declines \\
\midrule
\multirow{3}{*}{Qwen2.5 7B} & named & 60.4 & 17.9 & 21.7 \\
 & family-only & 35.3 & 32.6 & 32.1 \\
 & silent & 25.0 & 15.6 & 59.4 \\
\midrule
\multirow{3}{*}{OLMo 2 7B} & named & 65.1 & 31.1 & 3.8 \\
 & family-only & 36.7 & 56.7 & 6.5 \\
 & silent & 37.5 & 51.2 & 11.2 \\
\midrule
\multirow{3}{*}{Granite 3.1 8B} & named & 50.9 & 24.5 & 24.5 \\
 & family-only & 33.0 & 33.0 & 34.0 \\
 & silent & 26.2 & 26.2 & 47.5 \\
\midrule
\multirow{3}{*}{Qwen2.5 0.5B} & named & 39.6 & 56.6 & 3.8 \\
 & family-only & 30.2 & 69.8 & 0.0 \\
 & silent & 28.7 & 68.8 & 2.5 \\
\bottomrule
\end{tabular}
\caption{Closed-book outcomes (\%) on the 481 neighborhood-matched questions (106 named, 215 family-only, and 160 silent) when no passage is provided.}
\label{tab:closed_full}
\end{table}

%% file: tables/tab_all_page.tex
\begin{table}[ht]
\centering
\small
\setlength{\tabcolsep}{2.8pt}
\begin{tabular}{llccccc}
\toprule
Model & Passage & Corr. & Wrong & Decl. & Prec. & Conf. \\
\midrule
\multirow{3}{*}{Qwen2.5 7B} & named & 64.7 & 3.3 & 32.0 & 95.1 & 0.99 \\
 & family-only & 29.3 & 17.6 & 53.1 & 62.5 & 1.00 \\
 & silent & 13.0 & 2.7 & 84.3 & 83.0 & 0.99 \\
\midrule
\multirow{3}{*}{OLMo 2 7B} & named & 71.3 & 20.7 & 8.0 & 77.5 & 0.80 \\
 & family-only & 39.1 & 42.6 & 18.4 & 47.8 & 0.84 \\
 & silent & 32.0 & 28.0 & 40.0 & 53.3 & 0.76 \\
\midrule
\multirow{3}{*}{Granite 3.1 8B} & named & 64.3 & 4.7 & 31.0 & 93.2 & 0.67 \\
 & family-only & 33.2 & 25.0 & 41.8 & 57.0 & 0.93 \\
 & silent & 15.7 & 11.0 & 73.3 & 58.8 & 0.81 \\
\midrule
\multirow{3}{*}{Qwen2.5 0.5B} & named & 78.7 & 14.3 & 7.0 & 84.6 & 0.43 \\
 & family-only & 37.5 & 52.3 & 10.2 & 41.7 & 0.55 \\
 & silent & 28.7 & 50.0 & 21.3 & 36.4 & 0.39 \\
\bottomrule
\end{tabular}
\caption{Open-book outcomes across all 856 PrimeKG questions (\%; 300 named, 256 family-only, and 300 silent). Precision is the share of correct answers among non-abstained questions; Conf.\ is the median probability assigned to wrong answers.}
\label{tab:all_page}
\end{table}

%% file: tables/tab_did_full.tex
\begin{table*}[t]
\centering
\footnotesize
\setlength{\tabcolsep}{2.4pt}
\begin{tabular}{lccccccccc}
\toprule
 & \multicolumn{3}{c}{Declines (Abstention)} & \multicolumn{3}{c}{Correct Answers} & \multicolumn{3}{c}{Wrong Answers} \\
\cmidrule(lr){2-4}\cmidrule(lr){5-7}\cmidrule(lr){8-10}
Model & Fam. & Sil. & Diff.\ [95\% CI], $p$ & Fam. & Sil. & Diff.\ [95\% CI], $p$ & Fam. & Sil. & Diff.\ [95\% CI], $p$ \\
\midrule
Qwen2.5 7B & +23.3 & +24.4 & $-1.1$ [$-10.5$, 8.4], .82 & $-8.8$ & $-11.2$ & +2.4 [$-5.7$, 10.5], .57 & $-14.4$ & $-13.1$ & $-1.3$ [$-9.4$, 7.1], .76 \\
OLMo 2 7B & +11.6 & +33.8 & $-22.1$ [$-31.8$, $-12.6$], 6e-6 & +0.5 & $-6.9$ & +7.3 [$-2.6$, 17.0], .15 & $-12.1$ & $-26.9$ & +14.8 [3.6, 26.1], .009 \\
Granite 3.1 8B & +9.3 & +27.5 & $-18.2$ [$-28.5$, $-8.1$], .0006 & $-2.8$ & $-11.2$ & +8.5 [$-0.9$, 17.8], .064 & $-6.5$ & $-16.2$ & +9.7 [$-0.1$, 19.9], .056 \\
Qwen2.5 0.5B & +9.8 & +18.1 & $-8.4$ [$-15.7$, $-1.1$], .027 & +4.2 & 0.0 & +4.2 [$-6.1$, 14.4], .42 & $-14.0$ & $-18.1$ & +4.2 [$-6.4$, 14.6], .44 \\
\midrule
Pooled 7--8B & +14.7 & +28.5 & $-13.8$ [$-20.3$, $-7.5$], 4e-5 & $-3.7$ & $-9.8$ & +6.1 [$-0.1$, 12.4], .058 & $-11.0$ & $-18.8$ & +7.7 [1.2, 14.0], .019 \\
\bottomrule
\end{tabular}
\caption{Within-question page effects (in percentage points): the change in each outcome when adding a passage relative to closed-book answering on family-only (Fam., $n=215$) and silent (Sil., $n=160$) neighborhood-matched questions, along with the difference-in-differences estimate.}
\label{tab:did_full}
\end{table*}

%% file: tables/tab_byrel.tex
\begin{table}[ht]
\centering
\small
\setlength{\tabcolsep}{3pt}
\begin{tabular}{lcccc}
\toprule
 & \multicolumn{2}{c}{Indication} & \multicolumn{2}{c}{Contraindication} \\
\cmidrule(lr){2-3}\cmidrule(lr){4-5}
Model & Family-only & Silent & Family-only & Silent \\
\midrule
Qwen2.5 7B & 26.0 & 2.2 & 6.8 & 2.9 \\
OLMo 2 7B & 38.6 & 21.7 & 53.4 & 27.9 \\
Granite 3.1 8B & 37.8 & 16.3 & 10.2 & 1.5 \\
Qwen2.5 0.5B & 46.5 & 46.7 & 69.3 & 55.9 \\
\bottomrule
\end{tabular}
\caption{Open-book wrong-answer rates (\%) broken down by clinical relation on neighborhood-matched questions (indication: 127 family-only, 92 silent; contraindication: 88 family-only, 68 silent).}
\label{tab:byrel}
\end{table}

%% file: Supp_sections/E_edits.tex
\section{Page Edits}
\label{app:edits}

The abstention test compares different facts: the family-only questions are not the same facts as the silent ones. The page edits remove this difference. They ask each question three times, with three versions of one page, and change nothing else.

\paragraph{Which questions.}
We start from named questions, whose page names the answer. In PrimeKG we keep the 130 whose answer has a MONDO parent, since the family edit needs a parent name. In GloBI we keep the 255 named questions whose answer has a genus.

\paragraph{Three versions.}
\begin{itemize}
\item \emph{Original}: the page as retrieved. It names the answer.
\item \emph{Family}: every mention of the answer, and of each of its alternative names, becomes the name of its immediate parent (the MONDO parent, or the genus). In GloBI we also replace the abbreviated form (\emph{A.~plantaginifolia}).
\item \emph{Removed}: every sentence that mentions the answer is deleted.
\end{itemize}
We replace whole words only. If the answer has several parents, we use the shortest name, and we never use a parent that is itself one of the options. We drop a question if the answer is still mentioned after the edit, or if nothing is left after removal.
For example, a page that says ``\emph{\dots used to treat familial hyperlipidemia}'' becomes ``\emph{\dots used to treat hyperlipidemia}'' in the family version, and loses that sentence in the removed version. The question, the options, their letters and the model stay the same. Each prompt is cached under a hash of its exact text, so an edited page can never reuse the answer of the original.

\paragraph{What the comparison shows.}
The family and removed versions both lack the answer. The only difference is that the family version keeps the family name in place of the answer. If the family name had no effect, the two versions would give the same outcomes. \Cref{tab:edit_models} shows that they do not: in every model and both domains, the family version draws fewer declines than the removed version.

\begin{table}[t]
\centering
\footnotesize
\setlength{\tabcolsep}{2.6pt}
\begin{tabular}{@{}llrrrr@{}}
\toprule
Model & Version & Decl. & Corr. & Wrong & Prec. \\
\midrule
\multicolumn{6}{@{}l}{\emph{PrimeKG (130 questions)}} \\
Qwen2.5-7B & original & 27.7 & 66.9 & 5.4 & 92.6 \\
 & family & 43.8 & 46.2 & 10.0 & 82.2 \\
 & removed & 70.0 & 26.9 & 3.1 & 89.7 \\
OLMo 2 7B & original & 6.2 & 74.6 & 19.2 & 79.5 \\
 & family & 7.7 & 58.5 & 33.8 & 63.3 \\
 & removed & 18.5 & 52.3 & 29.2 & 64.2 \\
Granite 3.1 8B & original & 23.8 & 69.2 & 6.9 & 90.9 \\
 & family & 40.8 & 43.1 & 16.2 & 72.7 \\
 & removed & 56.2 & 31.5 & 12.3 & 71.9 \\
\midrule
\multicolumn{6}{@{}l}{\emph{GloBI (255 questions)}} \\
Qwen2.5-7B & original & 38.4 & 58.8 & 2.7 & 95.5 \\
 & family & 71.0 & 21.6 & 7.5 & 74.3 \\
 & removed & 92.2 & 5.1 & 2.7 & 65.0 \\
OLMo 2 7B & original & 17.3 & 71.4 & 11.4 & 86.3 \\
 & family & 25.9 & 44.3 & 29.8 & 59.8 \\
 & removed & 56.5 & 20.8 & 22.7 & 47.7 \\
Granite 3.1 8B & original & 32.2 & 61.2 & 6.7 & 90.2 \\
 & family & 50.2 & 32.9 & 16.9 & 66.1 \\
 & removed & 81.6 & 9.8 & 8.6 & 53.2 \\
\bottomrule
\end{tabular}
\caption{Outcomes (\%) with each version of the page. In every model and domain, the family version draws fewer declines than the removed version.}
\label{tab:edit_models}
\end{table}

\paragraph{Pooled effects.}
\Cref{tab:edit_pooled} pools the three models, averaging each question over models first. Against the removed version, the family name cuts declines by 17.4 points in PrimeKG and 27.7 in GloBI, and raises both wrong and correct answers. Against the original, the family version loses many correct answers and gains wrong ones. So the family name lets the model guess among the family members without telling it which one is meant.

\begin{table}[t]
\centering
\footnotesize
\setlength{\tabcolsep}{2.5pt}
\begin{tabular}{@{}llrr@{}}
\toprule
 & Outcome & Family $-$ removed & Family $-$ original \\
\midrule
\multirow{3}{*}{PrimeKG} & Declines & $-17.4$ [$-23.6$, $-11.5$] & $+11.5$ [$+7.4$, $+15.9$] \\
 & Wrong & $+5.1$ [$+0.5$, $+10.3$] & $+9.5$ [$+5.9$, $+13.6$] \\
 & Correct & $+12.3$ [$+6.9$, $+17.9$] & $-21.0$ [$-26.9$, $-15.6$] \\
\midrule
\multirow{3}{*}{GloBI} & Declines & $-27.7$ [$-32.2$, $-23.5$] & $+19.7$ [$+15.4$, $+23.9$] \\
 & Wrong & $+6.7$ [$+3.4$, $+10.2$] & $+11.1$ [$+8.2$, $+14.1$] \\
 & Correct & $+21.0$ [$+16.9$, $+25.4$] & $-30.8$ [$-36.2$, $-25.5$] \\
\bottomrule
\end{tabular}
\caption{Pooled edit effects over the three 7--8B models (points, 95\% bootstrap CI).}
\label{tab:edit_pooled}
\end{table}

%% file: Supp_sections/F_fixes.tex
\section{Three Fixes}
\label{app:fixes}

This appendix gives the per-model numbers behind \cref{fig:fixes} of the main paper.

\paragraph{Confidence.}
A simple guard is to answer only when the model is confident. \Cref{tab:conf} shows, for each model and page type, how many questions the model answers, its precision, its mean confidence on right and on wrong answers, and the AUROC (\cref{app:stats}). On named pages, right answers carry clearly more confidence than wrong ones, and AUROC is 0.73--0.87. On family-only pages the two are close, and AUROC drops to 0.59--0.72. The wrong guesses are confident: Qwen2.5-14B puts a mean probability of 0.99 on its wrong family-only answers. In GloBI the drop is larger, from 0.82--0.86 on named pages to 0.52--0.68.

\begin{table}[t]
\centering
\footnotesize
\setlength{\tabcolsep}{2.2pt}
\begin{tabular}{@{}llrrrrr@{}}
\toprule
Model & Page & Ans. & Prec. & Conf.\,R & Conf.\,W & AUROC \\
\midrule
\multicolumn{7}{@{}l}{\emph{PrimeKG}} \\
Qwen2.5-0.5B & named & 97 & 84.5 & .83 & .49 & .85 \\
 & family & 194 & 38.1 & .71 & .60 & .64 \\
Qwen2.5-7B & named & 80 & 95.0 & .98 & 1.00 & .73 \\
 & family & 96 & 59.4 & .98 & .90 & .72 \\
OLMo 2 7B & named & 98 & 83.7 & .97 & .75 & .86 \\
 & family & 176 & 45.5 & .89 & .77 & .71 \\
Granite 3.1 8B & named & 85 & 90.6 & .95 & .71 & .87 \\
 & family & 122 & 53.3 & .90 & .82 & .66 \\
Qwen2.5-14B & named & 75 & 96.0 & .99 & .92 & .77 \\
 & family & 89 & 58.4 & .98 & .99 & .59 \\
\midrule
\multicolumn{7}{@{}l}{\emph{GloBI}} \\
Qwen2.5-7B & named & 167 & 94.6 & .98 & .90 & .82 \\
 & family & 18 & 27.8 & .94 & .89 & .55 \\
OLMo 2 7B & named & 231 & 84.4 & .94 & .69 & .86 \\
 & family & 138 & 26.8 & .76 & .62 & .68 \\
Granite 3.1 8B & named & 188 & 89.4 & .94 & .75 & .86 \\
 & family & 46 & 21.7 & .75 & .74 & .52 \\
\bottomrule
\end{tabular}
\caption{Confidence of answers with the page. Ans.\ is the number of questions answered (not declined); Conf.\,R and Conf.\,W are the mean confidence of right and wrong answers.}
\label{tab:conf}
\end{table}

\paragraph{Scale.}
Within Qwen2.5, a larger model is more precise on named pages: 84.5\% at 0.5B, 95.0\% at 7B and 96.0\% at 14B. On family-only pages, precision rises from 38.1\% to 59.4\% between 0.5B and 7B and then stops: the 14B model reaches 58.4\%. The 14B model gives 14.7 points more wrong answers on family-only pages than on silent ones ($p=6\times10^{-6}$). Like the 7B model, it already declines family-only questions less often closed book, so its page effect does not differ between the groups ($+5.3$ points, $p=0.29$). A larger model reads named pages better but still guesses from a family name.

\paragraph{Instruction.}
We add one sentence to the prompt with the page:
\begin{quote}
\small\emph{If the passage names only a broad category of diseases and does not say which specific disease is meant, answer with the letter of ``Not enough information to tell.}
\end{quote} \Cref{tab:instr} shows the result. The models react very differently. OLMo~2 declines far more on every page type, including named pages (from 7.5\% to 27.4\%). Granite barely reacts. Qwen2.5-7B declines somewhat more. Pooled over the three models, the instruction raises declines by 15.2 points on family-only pages, but also by 14.0 on silent pages and 9.1 on named pages, where it costs 6.3 points of correct answers. Precision on family-only pages stays low (47.9--64.9\%). The instruction makes the models more cautious everywhere, but it does not teach them to tell a family name from an answer.

\begin{table}[t]
\centering
\footnotesize
\setlength{\tabcolsep}{2.4pt}
\begin{tabular}{@{}llrrrr@{}}
\toprule
 & & \multicolumn{2}{c}{Declines} & \multicolumn{2}{c}{Precision} \\
\cmidrule(lr){3-4}\cmidrule(l){5-6}
Model & Page & base & instr. & base & instr. \\
\midrule
Qwen2.5-7B & named & 24.5 & 29.2 & 95.0 & 94.7 \\
 & family & 55.4 & 64.2 & 59.4 & 64.9 \\
 & silent & 83.8 & 88.1 & 84.6 & 89.5 \\
OLMo 2 7B & named & 7.5 & 27.4 & 83.7 & 89.6 \\
 & family & 18.1 & 56.3 & 45.5 & 47.9 \\
 & silent & 45.0 & 85.6 & 55.7 & 60.9 \\
Granite 3.1 8B & named & 19.8 & 22.6 & 90.6 & 91.5 \\
 & family & 43.3 & 41.9 & 53.3 & 53.6 \\
 & silent & 75.0 & 71.9 & 60.0 & 55.6 \\
\bottomrule
\end{tabular}
\caption{Declines and precision (\%) with the page, without (base) and with (instr.) the instruction to decline on a broad category.}
\label{tab:instr}
\end{table}

%% file: Supp_sections/G_globi.tex
\section{Second Domain: GloBI}
\label{app:globi}

We repeat the abstention test and the page edits on species interactions. This tests whether \clade{} depends on medicine, on MONDO, or on PrimeKG. It does not: the same pattern appears with a different graph, a different hierarchy and different text.

\subsection{Setup}

\paragraph{Facts and pages.}
Each fact links a consumer species to a partner species, such as a butterfly and the plant its larvae eat. The page is the consumer's Wikipedia article, cleaned as in \cref{app:data}. A page is \emph{named} if it gives the partner's Latin name, its abbreviated form (\emph{A.~plantaginifolia}) or a common name from Wikipedia redirects. It is \emph{family-only} if it gives only the partner's genus or family, and \emph{silent} if it gives neither. We measure \num{23694} edges over \num{5425} articles. Pages name the partner for only 3.9\% of edges and name only its genus or family for 7.1\% (\cref{tab:coverage}). Family-only pages are thus about twice as common as named ones.

\paragraph{Questions.}
We use one template per relation:
\begin{center}
\footnotesize
\begin{tabular}{@{}ll@{}}
\toprule
Relation & Question \\
\midrule
eats & Which species does X feed on? \\
preys on & Which species does X prey on? \\
parasite of & Which species is X a parasite of? \\
pollinates & Which plant does X pollinate? \\
\bottomrule
\end{tabular}
\end{center}
We take 75 questions per page type and relation, for 900 in all. Wrong options are other species of the partner's genus first, then of its family, and never another recorded partner of the consumer. For binomial answers, we fill any remaining slots with species from other genera, never with a bare genus name, so every option is a species. Most questions have four neighborhood options (833 of 900; \cref{tab:globi_sib}), and the main analysis keeps only these.

\begin{table}[t]
\centering
\footnotesize
\setlength{\tabcolsep}{3.5pt}
\begin{tabular}{@{}lrrrr@{}}
\toprule
Page type & eats & parasite of & pollinates & preys on \\
\midrule
Named & 93.3 & 94.7 & 94.7 & 89.3 \\
Family-only & 100.0 & 98.7 & 100.0 & 93.3 \\
Silent & 90.7 & 85.3 & 92.0 & 78.7 \\
\bottomrule
\end{tabular}
\caption{Share of GloBI questions (\%, 75 per cell) whose four wrong options are all from the partner's genus or family.}
\label{tab:globi_sib}
\end{table}

\subsection{Results per Model}

\paragraph{Outcomes.}
\Cref{tab:globi_models} gives outcomes for each model, closed book and with the page. Closed book, all three models decline most species questions; they know few of these facts. A named page lets them answer, and their answers are precise (84--95\%). A family-only page also lets them answer, but most of these answers are wrong: precision is 22--28\%, close to the 20\% chance rate of a five-way guess. A silent page makes them decline almost everything.

\begin{table}[t]
\centering
\footnotesize
\setlength{\tabcolsep}{2.4pt}
\begin{tabular}{@{}llrrrrrr@{}}
\toprule
 & & \multicolumn{2}{c}{Correct} & \multicolumn{2}{c}{Wrong} & \multicolumn{2}{c}{Declines} \\
\cmidrule(lr){3-4}\cmidrule(lr){5-6}\cmidrule(l){7-8}
Model & Page & cl. & page & cl. & page & cl. & page \\
\midrule
Qwen2.5-7B & named & 7.9 & 56.6 & 4.3 & 3.2 & 87.8 & 40.1 \\
 & family & 1.7 & 1.7 & 2.7 & 4.4 & 95.6 & 93.9 \\
 & silent & 3.5 & 0.4 & 1.2 & 1.2 & 95.4 & 98.5 \\
OLMo 2 7B & named & 20.8 & 69.9 & 24.7 & 12.9 & 54.5 & 17.2 \\
 & family & 10.9 & 12.6 & 22.8 & 34.4 & 66.3 & 53.1 \\
 & silent & 8.5 & 3.5 & 22.3 & 8.8 & 69.2 & 87.7 \\
Granite 3.1 8B & named & 12.2 & 60.2 & 16.8 & 7.2 & 71.0 & 32.6 \\
 & family & 9.2 & 3.4 & 14.6 & 12.2 & 76.2 & 84.4 \\
 & silent & 5.8 & 1.5 & 12.3 & 1.9 & 81.9 & 96.5 \\
\bottomrule
\end{tabular}
\caption{GloBI outcomes (\%) closed book (cl.) and with the page, on questions whose four wrong options all come from the neighborhood (279 named, 294 family-only, 260 silent).}
\label{tab:globi_models}
\end{table}

\paragraph{Page effects.}
\Cref{tab:globi_did} gives the page effect for each model. The effect is largest in OLMo~2, where a silent page raises declines by 18.5 points but a family-only page lowers them by 13.3, a difference of 31.7 points. Qwen2.5-7B declines almost every question on either page, so its effect is small but still significant. Granite's decline effect is not significant alone, but its wrong answers rise reliably. Pooled over the three models, $\Delta=-14.3$ points for declines and $+11.6$ for wrong answers, with only $+2.7$ for correct answers: in GloBI, the answers that the family name frees are mostly wrong.

\begin{table}[t]
\centering
\footnotesize
\setlength{\tabcolsep}{2.4pt}
\begin{tabular}{@{}llrlr@{}}
\toprule
Model & Outcome & $\Delta$ & [95\% CI] & $p$ \\
\midrule
Qwen2.5-7B & declines & $-4.8$ & [$-8.9$, $-0.8$] & .022 \\
 & wrong & $+1.7$ & [$-1.1$, $+4.8$] & .26 \\
OLMo 2 7B & declines & $-31.7$ & [$-40.6$, $-22.7$] & 2e-11 \\
 & wrong & $+25.0$ & [$+16.5$, $+33.4$] & 1e-8 \\
Granite 3.1 8B & declines & $-6.5$ & [$-14.0$, $+0.8$] & .097 \\
 & wrong & $+8.0$ & [$+1.5$, $+14.6$] & .019 \\
\midrule
Pooled & declines & $-14.3$ & [$-18.7$, $-9.6$] & 1e-9 \\
 & wrong & $+11.6$ & [$+7.5$, $+15.4$] & 3e-8 \\
\bottomrule
\end{tabular}
\caption{GloBI page effects (points), family-only minus silent (Eq.~\ref{eq:did}).}
\label{tab:globi_did}
\end{table}

\paragraph{Page edits and confidence.}
The per-model edit results are in \cref{tab:edit_models} and the confidence results in \cref{tab:conf}. In every model, the genus name in place of the answer draws fewer declines than deleting the sentence, and AUROC on family-only pages falls to 0.52--0.68.

\paragraph{Why GloBI is harder.}
Sibling species are harder to tell apart than sibling diseases. The models know few species facts closed book (\cref{tab:globi_models}), so when a page names only the genus, they guess inside the genus close to chance. This is why precision on family-only pages is near chance in GloBI (22--28\%) but closer to half in PrimeKG (45--59\%).

%% file: Supp_sections/H_recovery.tex
\section{Ontology Recovery}
\label{app:recovery}

Recovery asks whether the ontology can turn a family name back into the exact answer. It has three steps: build a list of candidates, choose one, and compare with baselines that do not use the ontology.

\paragraph{Facts.}
We use all 410 PrimeKG indications and contraindications whose combined Wikipedia and PubMed text names only a MONDO family that expands to at least two diseases. A drug often has several true diseases in the same family (2.47 per list on average). We therefore count a pick as correct if it matches any disease the drug has under that relation. \Cref{tab:scoring} also gives the strict score, which accepts only the one sampled answer.

\subsection{Building the List}

\paragraph{Given family.}
This list assumes we already know which family on the page belongs to the question. We start at the true disease and climb up to two levels in MONDO. Among these ancestors, we keep the ones the drug's text names. For each, we collect its descendants up to two levels down. We keep the ancestor with the smallest list, sort the list, and cap it at 64 diseases. This is an oracle: it uses the true disease to pick the family. It tells us how good the list can be at best.

\paragraph{Page built.}
This list uses only the page, as a real system would. We take all 3,182 MONDO families with two or more children as candidate family names. We index every run of up to eight words on the drug's cleaned Wikipedia page, adding singular forms of words that end in \emph{s}. Every family whose name appears in this index is a match. We drop any matched family whose own two-level list exceeds 64 diseases, and join the lists of the rest. If the joined list still exceeds 64, we mark the page as \emph{too broad}. In short:
\begin{quote}
\footnotesize\ttfamily
fams = [f for f in families if name(f) in page]\\
fams = [f for f in fams if |list(f)| <= 64]\\
cands = union(list(f) for f in fams)\\
if |cands| > 64: too broad
\end{quote}

\paragraph{Why page-built lists fail.}
A drug article names many disease families that are not the answer's: other uses, side effects, warnings and conditions that often occur together. The median Wikipedia page matches five families; with PubMed it matches 17. Their union is usually too large (\cref{tab:pagebuilt}). The family that holds the answer is on the page, but the matcher cannot tell it apart from the others.

\begin{table}[t]
\centering
\footnotesize
\setlength{\tabcolsep}{3pt}
\begin{tabular}{@{}lrr@{}}
\toprule
 & Wikipedia & + PubMed \\
\midrule
Matched families per page (median) & 5 & 17 \\
Usable lists (of 410) & 166 & 19 \\
Lists over the 64 cap & 209 & 391 \\
Usable list holds sampled answer & 19.8\% & 3.7\% \\
Usable list holds any true answer & 24.6\% & 4.1\% \\
Median usable list size & 32 & 59 \\
End-to-end recovery (evidence pick) & 10.0\% & 1.5\% \\
\bottomrule
\end{tabular}
\caption{Page-built lists from Wikipedia alone and from Wikipedia plus PubMed, over all 410 family-only facts. The remaining unusable lists match no family with two or more children.}
\label{tab:pagebuilt}
\end{table}

\subsection{Choosing One Candidate}

\paragraph{Model scores.}
Qwen2.5-7B scores each candidate by its mean log-probability as the completion of a relation stem, such as ``\emph{X is a drug used to treat the disease called}''. In the \emph{model alone} condition the prompt has only the stem; in the \emph{model with page} condition the page comes first. The highest score is the pick.

\paragraph{Evidence scores.}
We use two outside sources. The \emph{Wikipedia reverse lookup} checks whether the candidate disease's own article names the drug (a \emph{mention}), and whether a relation keyword appears within 250 characters of that mention (a \emph{cue}). Indication keywords are \emph{treat}, \emph{therap}, \emph{indicated} and \emph{first-line}; contraindication keywords are \emph{contraindicat}, \emph{avoid}, \emph{must not}, \emph{caution}, \emph{worsen} and \emph{adverse}. The \emph{PubMed co-mention} count is the number of records that name both the drug and the candidate with a relation qualifier:
\begin{quote}
\footnotesize\ttfamily
"<drug>"[tiab] AND "<cand>"[tiab] AND <q>
\end{quote}
where the indication qualifier \texttt{<q>} is \texttt{(treat*[tiab] OR therap*[tiab] OR efficac*[tiab])} and the contraindication qualifier is \texttt{(contraindicat*[tiab] OR adverse[tiab] OR toxicit*[tiab] OR "side effect"[tiab] OR "side effects"[tiab] OR induced[tiab])}. We strip parenthetical parts such as ``(disease)'' from MONDO labels before we query. The combined evidence score of candidate $c$ is
\begin{equation}
E(c)=2\cdot\mathbf{1}[\text{cue}]+\mathbf{1}[\text{mention}]+\log\bigl(1+\text{co-mentions}\bigr).
\end{equation}
A signal makes a pick only when one candidate has the strictly highest score; otherwise it records a tie or no signal. The \emph{evidence, then model} chooser uses $E$ when it makes a pick and falls back to the model otherwise.

\paragraph{Popularity prior.}
The popularity prior ignores the drug. It counts all PubMed records that name the candidate (\texttt{"<cand>"[tiab]}) and picks the candidate with the most. The \emph{normalized} co-mention signal divides each co-mention count by this total, which removes the popularity part.

\paragraph{What each signal does.}
\Cref{tab:signals} shows how often each signal makes a pick and how accurate those picks are. The Wikipedia lookup is accurate when it fires (50.0\%), but it fires for only 82 facts, because a disease article rarely names a specific drug. PubMed co-mentions fire often, but they are only as accurate as the popularity prior. Once we divide out popularity, accuracy falls from 35.9\% to 25.4\%. Most of the signal in co-mentions is therefore how much a disease is studied, not its link to the drug.

\begin{table}[t]
\centering
\footnotesize
\setlength{\tabcolsep}{2.2pt}
\begin{tabular}{@{}lrrrrr@{}}
\toprule
Signal & Picks & Ties & None & Acc./pick & Acc./all \\
\midrule
Wikipedia lookup & 82 & 80 & 248 & 50.0 & 10.0 \\
PubMed co-mention & 324 & 12 & 74 & 45.4 & 35.9 \\
Combined evidence & 327 & 14 & 69 & 44.6 & 35.6 \\
Popularity prior & 409 & 0 & 1 & 46.9 & 46.8 \\
Normalized PubMed & 336 & 0 & 74 & 31.0 & 25.4 \\
\bottomrule
\end{tabular}
\caption{Signals inside the given-family list (410 facts, \num{8448} candidates). Accuracy (\%) counts a pick as correct if it matches any true disease of the drug.}
\label{tab:signals}
\end{table}

\subsection{Baselines Without the Ontology}

To see what the ontology adds, we replace the family list with a list of the same length that holds the sampled answer and random diseases from the same relation (never another true partner of the drug). The model scores this \emph{random list} with or without the page. \Cref{tab:scoring} compares the conditions. Under the strict score, the random list looks easier, because its wrong candidates are unrelated diseases that the model can rule out. Under the any-true-answer score, the family list catches the drug's other true diseases, and the model does about as well on it as on the random list. The family list is valuable because it is short and complete (98.5\% coverage), not because it makes the model's choice easier.

\begin{table}[t]
\centering
\footnotesize
\setlength{\tabcolsep}{3pt}
\begin{tabular}{@{}lrr@{}}
\toprule
Condition & Sampled & Any true \\
\midrule
Model, given-family list & 18.3 & 35.9 \\
Model + page, given-family list & 17.8 & 32.4 \\
Model, random list & 30.5 & 34.9 \\
Model + page, random list & 29.0 & 32.7 \\
\bottomrule
\end{tabular}
\caption{Qwen2.5-7B recovery accuracy (\%) on the 410 family-only facts, scoring only the sampled answer (Sampled) or any true answer of the drug (Any true).}
\label{tab:scoring}
\end{table}

\subsection{Examples}

\Cref{tab:examples} shows evidence picks within the given-family list. When the pick misses the PrimeKG answer, it usually takes a better-studied member of the same family (\emph{ovarian carcinoma} for carboplatin) or a true use that PrimeKG does not list (\emph{Legionnaires' disease} for moxifloxacin, which is used against this pneumonia). Strict scoring against the graph counts both as wrong, so it understates how often the pick is clinically sound.

\input{tables/tab_examples}

%% file: tables/tab_examples.tex
\begin{table}[ht]
\centering
\small
\setlength{\tabcolsep}{3pt}
\begin{tabular}{>{\raggedright\arraybackslash}p{1.85cm}>{\raggedright\arraybackslash}p{2.35cm}>{\raggedright\arraybackslash}p{2.2cm}c}
\toprule
Drug & PrimeKG Target & Evidence Pick & Match? \\
\midrule
Lovastatin & familial hyperlipidemia & familial hyperlipidemia & yes \\
Tetrabenazine & juvenile Huntington disease & juvenile Huntington disease & yes \\
Moxifloxacin & Klebsiella pneumonia & Legionnaires' disease & no \\
Carboplatin & malignant germ cell tumor of ovary & ovarian carcinoma & no \\
\bottomrule
\end{tabular}
\caption{Qualitative examples of candidate diseases chosen by combined literature evidence within the given-family list on indication facts.}
\label{tab:examples}
\end{table}

%% file: Supp_sections/I_cases.tex
\section{Worked Examples}
\label{app:cases}

This appendix follows single questions through the test. Each example shows the page, the question and what each model answers, closed book and with the page. Page excerpts come from English Wikipedia (CC BY-SA 4.0); ``\dots'' marks text we leave out.

\newcommand{\casebox}[1]{\leavevmode\unskip\hspace*{\fill}\par\noindent\fbox{\parbox{\dimexpr\columnwidth-2\fboxsep-2\fboxrule\relax}{\footnotesize #1}}\par\medskip}

\subsection{PrimeKG}

\paragraph{Moxifloxacin: a confident wrong guess.}
\casebox{\textbf{Fact:} moxifloxacin treats \emph{Klebsiella pneumonia}.\\
\textbf{Page type:} family-only. The page says the drug treats \emph{pneumonia}, but not which kind.\\
\textbf{Question:} Which disease is moxifloxacin used to treat?\\
\textbf{Options include:} Klebsiella pneumonia, Legionnaires' disease, pneumonia caused by gram negative bacteria, and other bacterial pneumonias; Not enough information to tell.}

Every model answers, and every model is wrong:

\begin{center}
\footnotesize
\setlength{\tabcolsep}{3pt}
\begin{tabular}{@{}lll@{}}
\toprule
Model & Closed book & With the page (conf.) \\
\midrule
Qwen2.5-0.5B & granegative pn. & granegative pn. (.35) \\
Qwen2.5-7B & Legionnaires' & Legionnaires' (.88) \\
OLMo 2 7B & granegative pn. & granegative pn. (.86) \\
Granite 3.1 8B & granegative pn. & Legionnaires' (.77) \\
Qwen2.5-14B & Legionnaires' & granegative pn. (1.00) \\
\bottomrule
\end{tabular}
\end{center}

The page tells the models that the answer is a pneumonia, and they pick a pneumonia with high confidence. None of them declines, even though the page cannot say which pneumonia is meant. A confidence threshold would keep most of these answers. Note also that \emph{Legionnaires' disease} is a real use of moxifloxacin that PrimeKG does not list: the graph counts a sound answer as wrong (\cref{app:recovery}).

\paragraph{Lovastatin: the page makes one model worse.}
\casebox{\textbf{Fact:} lovastatin treats \emph{hyperlipidemia, familial combined, LPL related}.\\
\textbf{Page type:} family-only. The page says \emph{hyperlipidemia}.\\
\textbf{Family in MONDO:} hyperlipidemia $\rightarrow$ familial hyperlipidemia $\rightarrow$ \{the answer, hyperlipidemia due to hepatic triglyceride lipase deficiency, familial lipoprotein lipase deficiency, \dots\}.}

\begin{center}
\footnotesize
\setlength{\tabcolsep}{3pt}
\begin{tabular}{@{}lll@{}}
\toprule
Model & Closed book & With the page \\
\midrule
Qwen2.5-0.5B & wrong sibling & wrong sibling \\
Qwen2.5-7B & declines & \textbf{correct} (1.00) \\
OLMo 2 7B & \textbf{correct} (.95) & wrong sibling (.54) \\
Granite 3.1 8B & \textbf{correct} (.98) & \textbf{correct} (.86) \\
Qwen2.5-14B & wrong sibling & wrong sibling (1.00) \\
\bottomrule
\end{tabular}
\end{center}

Here the family name helps Qwen2.5-7B and hurts OLMo~2. OLMo~2 knows the answer closed book, but with the page it switches to \emph{hyperlipidemia due to hepatic triglyceride lipase deficiency}, a sibling. The page points to the right family, and the model then picks the wrong member. This is \clade{} in its plainest form.

\subsection{GloBI}

\paragraph{Named page: the answer is on the page.}
\casebox{\textbf{Page} (\emph{Congriscus megastomus}): ``\emph{Congriscus megastomus is an eel in the family Congridae (conger/garden eels). \dots\ It is a marine, temperate-water dwelling eel which is known from Japan and the Kyushu--Palau Ridge \dots}'' The page later names the prey.\\
\textbf{Question:} Which species does Congriscus megastomus feed on?\\
\textbf{Options:} Ventrifossa garmani, Ventrifossa mucocephalus, Coryphaenoides zaniophorus, \emph{Hymenocephalus lethonemus} (true), Gadomus colletti, Not enough information to tell.}

All three models decline closed book. With the page, Qwen2.5-7B and Granite pick the true species (confidence 1.00 and .997); OLMo~2 declines. When the page names the answer, the models use it well.

\paragraph{Family-only page: the page turns declines into guesses.}
\casebox{\textbf{Page} (\emph{Vanessa virginiensis}, American painted lady): ``\emph{The larvae feed on various Asteraceae, such as the cudweeds (genus Gnaphalium), the pussytoes (\textbf{Antennaria}), and the everlastings (Anaphalis) \dots}''\\
\textbf{Question:} Which species does Vanessa virginiensis feed on?\\
\textbf{Options:} Antennaria pulcherrima, \emph{Antennaria plantaginifolia} (true), Antennaria dimorpha, Antennaria lanata, Antennaria microphylla, Not enough information to tell.}

Every option is an \emph{Antennaria} species, so the genus name on the page rules out none of them. Closed book, all three models decline. With the page, Qwen2.5-7B still declines, but OLMo~2 picks \emph{A.~microphylla} (.52) and Granite picks \emph{A.~pulcherrima} (.63). Both are wrong. The page did not tell them the answer; it only made the question look answerable.

\paragraph{Silent page: the models abstain.}
\casebox{\textbf{Page} (\emph{Cottunculus microps}, polar sculpin): ``\emph{\dots a species of fathead sculpin, a deepwater fish found in the North Atlantic and Arctic Oceans. \dots\ Like other members of its family, the polar sculpin has a large head, a short tapering body and a small tail. \dots}''\\
\textbf{Question:} Which species does Cottunculus microps feed on?\\
\textbf{True answer:} \emph{Lumpenus lampretaeformis}.}

The page describes the fish but says nothing about its food. All three models decline, closed book and with the page. This is the behavior we want on a family-only page as well, unless the system hands the question to the taxonomy.

\paragraph{A named page the models miss.}
The page for \emph{Orgyia vetusta} (western tussock moth) names its food plant only by the common name \emph{toyon}, which redirects to \emph{Photinia arbutifolia}. Our matcher counts this page as named. All three models decline anyway: they do not link the common name to the Latin name in the options. Common names are a second gap between how text names a fact and how a graph stores it.

%% file: Supp_sections/J_compute.tex
\section{Software and Compute}
\label{app:compute}

\paragraph{Models.}
We use the following instruction-tuned checkpoints from Hugging Face: {\small\textit{Qwen2.5-0.5B-Instruct}}, {\small\textit{Qwen2.5-7B-Instruct}}, {\small\textit{Qwen2.5-14B-Instruct}}, {\small\textit{OLMo-2-1124-7B-Instruct}}, {\small\textit{granite-3.1-8b-instruct}}. We do not fine-tune any model.

\paragraph{Running the models.}
All models run in 32-bit floating point with Hugging Face Transformers and PyTorch, on CPU nodes of a university cluster. Each model answers all 1,712 abstention prompts (856 questions, closed book and with the page) in one 12-thread job. We score prompts in batches with left padding. We rebuild the position indices from the attention mask, so a batched prompt gets exactly the same probabilities as the same prompt alone. Every answer is stored under a hash of its exact prompt, so an edited page never reuses an old answer.

\paragraph{Recovery.}
In recovery, all candidates for one fact share the same prompt prefix. We encode this prefix once and reuse its key--value cache for every candidate. Before each run, we check a sample of these cached scores against full forward passes.

\paragraph{Randomness and statistics.}
All sampling uses seed 0, and the bootstrap draws are sorted by question identifier before resampling, so every interval in the paper is the same on every run. The statistics use only the Python standard library.

%% file: Supp_sections/K_followup.tex
\section{Choosing the Family and Free-Text Answers}
\label{app:followup}

This appendix covers two follow-up tests. The first tests a simple fix for step~2 of the hand-off: which family on the page holds the answer. The second asks the abstention questions without options, to check that the option list does not drive the main result.

\subsection{Choosing the Family on the Page}

\paragraph{Setup.}
We use the same 410 family-only facts as in \cref{app:recovery}. For each fact, we take the families that the page-built list finds on the drug's Wikipedia page and keep those whose own list has at most 64 diseases. We order them by where the page first mentions them and show at most ten, labeled A to J. Qwen2.5-7B reads a window of the page around these mentions and the question, and picks one family:
\begin{quote}
\footnotesize\ttfamily
Passage:\\
<page window>\\[2pt]
Question: Which disease is X used to treat?\\
The answer is one specific disease. Which of these disease categories from the passage contains it?\\
Options:\\
A. <family>\\
\ldots\\
Answer with the letter of one option only.
\end{quote}
We read the answer from the letter probabilities, as in \cref{app:questions}. We then expand only the chosen family, two levels down, and score its members with the same choosers as before.

\paragraph{Rules that read no meaning.}
To see what the model adds, we compare it with three rules that ignore the question: the family the page mentions first, the family with the smallest list, and a random family. We also report a ceiling: the share of facts where at least one of the shown families holds the answer.

\paragraph{Results.}
\Cref{tab:pick} shows the result. The model's choice holds the answer for 30.0\% of all 410 facts, against 19.8\% for the page-built list, which joins every family on the page. It also beats each rule. The first-mentioned family is a strong rule (25.1\%); the smallest family is a poor one (7.8\%), worse than a random pick. The ceiling is 50.0\%: for the other half of the facts, the family that holds the answer is not among the usable families shown. The model therefore closes 30.0 of the 50.0 points that any chooser over these families could reach.

\begin{table}[t]
\centering
\footnotesize
\setlength{\tabcolsep}{3pt}
\begin{tabular}{@{}lrr@{}}
\toprule
Rule & Sampled answer & Any true answer \\
\midrule
Qwen2.5-7B picks the family & \textbf{30.0} & \textbf{44.9} \\
First mentioned & 25.1 & 37.3 \\
Random & 13.9 & 22.0 \\
Smallest family & 7.8 & 13.2 \\
\midrule
Page-built list (all families) & 19.8 & 24.6 \\
Any shown family (ceiling) & 50.0 & 64.6 \\
\bottomrule
\end{tabular}
\caption{Share of the 410 family-only facts (\%) whose chosen list holds the sampled answer, or any true answer of the drug. At most ten families are shown per page.}
\label{tab:pick}
\end{table}

\paragraph{End to end.}
The chosen lists are short (median 12 diseases), and 375 of the 410 facts get one. Scored with the same choosers, they recover the answer for 19.3\% of facts with the combined evidence and 26.8\% with the popularity prior. The page-built lists reach 10.0\% and 9.0\%. With no oracle at any step, a single prompted choice of the family nearly triples end-to-end recovery.

\subsection{Free-Text Answers}

\paragraph{Setup.}
We sample 40 questions per page type (named, family-only, silent) from the 481 questions whose four wrong options all come from the answer's neighborhood, with seed 0. We ask each one closed book and with the page, with no options:
\begin{quote}
\footnotesize\ttfamily
Passage:\\
<page>\\[2pt]
Question: Which disease is X used to treat?\\
Answer with the name of one disease only. If the passage does not tell you which one, answer "Not enough information to tell".
\end{quote}
Each model writes up to 32 tokens with greedy decoding.

\paragraph{Scoring.}
We check each answer in this order and assign the first outcome that applies:
\begin{enumerate}
\item \emph{correct}: it names the answer or one of its alternative names;
\item \emph{abstain}: it says there is not enough information, or that the passage does not say;
\item \emph{sibling}: it names one of the four wrong options of the multiple-choice question;
\item \emph{family}: it names one of the answer's family names, such as \emph{acute myeloid leukemia} for \emph{acute myeloid leukemia with t(8;21)};
\item \emph{other}: it names any other disease.
\end{enumerate}
The scoring is lenient: an answer that lists several diseases counts as correct if one of them is the answer. This affects only Granite on named pages, where 7 of its 9 correct answers are lists. It does not affect family-only or silent pages, where no model is ever correct.

\begin{table}[t]
\centering
\footnotesize
\setlength{\tabcolsep}{2.4pt}
\begin{tabular}{@{}llrrrrr@{}}
\toprule
Model & Page & Corr. & Abst. & Family & Sibl. & Other \\
\midrule
Qwen2.5-7B & named & 20.0 & 47.5 & 0.0 & 0.0 & 32.5 \\
 & family & 0.0 & 42.5 & 17.5 & 0.0 & 40.0 \\
 & silent & 0.0 & 52.5 & 0.0 & 0.0 & 47.5 \\
OLMo 2 7B & named & 25.0 & 37.5 & 0.0 & 0.0 & 37.5 \\
 & family & 0.0 & 32.5 & 20.0 & 2.5 & 45.0 \\
 & silent & 0.0 & 52.5 & 0.0 & 0.0 & 47.5 \\
Granite 3.1 8B & named & 22.5 & 37.5 & 0.0 & 0.0 & 40.0 \\
 & family & 0.0 & 30.0 & 45.0 & 2.5 & 22.5 \\
 & silent & 0.0 & 45.0 & 0.0 & 0.0 & 55.0 \\
\bottomrule
\end{tabular}
\caption{Free-text outcomes with the page (\%, 40 questions per row).}
\label{tab:free}
\end{table}

\paragraph{Results.}
\Cref{tab:free} gives the outcomes with the page. Three patterns stand out.
\begin{itemize}
\item \emph{No exact answers from family-only pages.} No model ever names the exact answer from a family-only or silent page. When the page names the answer, the models name it 20.0--25.0\% of the time.
\item \emph{Answering with the family.} On family-only pages, 17.5--45.0\% of answers name the family itself, such as ``acute myeloid leukemia (AML)''. Closed book, this happens 10.0--15.0\% of the time. It never happens on named or silent pages. Without options, \clade{} takes this form: the model does not pick a sibling, but gives the coarse category as if it were the answer.
\item \emph{Fewer abstentions.} Compared with silent pages, family-only pages lower abstention in all three models. The page effects (Eq.~\ref{eq:did}) are $-5.0$ [$-25.0$, $+15.0$] for Qwen2.5-7B, $-17.5$ [$-45.0$, $+7.5$] for OLMo~2 and $-10.0$ [$-32.5$, $+10.0$] for Granite. All three point the same way as in the main test, but with 40 questions per group no single interval excludes zero.
\end{itemize}
Free text also makes the models more cautious overall. They abstain on 37.5--47.5\% of named pages, where the multiple-choice test shows 7.5--24.5\%. Writing the answer is harder than picking a letter, so we compare page types within the free-text format, not across formats.